%% file: main.tex
\documentclass{article} 
\usepackage{iclr2027_conference,times}

\input{math_commands.tex}

\usepackage{hyperref}
\usepackage{url}
\usepackage{graphicx}
\usepackage{booktabs}
\usepackage{makecell}
\usepackage{amsmath}
\usepackage{amssymb}
\usepackage{tikz}
\usetikzlibrary{arrows.meta,positioning,fit,backgrounds,calc}

\newcommand{\pp}{\,pp}

\title{Published Unlearning Numbers Move Per Checkpoint, and Not Because the Removed Data Survives: An Audit of 263 Released Batch-Normalized Checkpoints}

\author{Junlong Shen \& Xingyu Li \\
University of Alberta \\
Edmonton, Canada \\
\texttt{junlong6@ualberta.ca}}

\iclrfinalcopy

\begin{document}

\maketitle
\lhead{Preprint}

\begin{abstract}
An unlearning audit reads its verdict off numbers that an unlearned model and its retrained
reference each publish, and both also ship batch-normalization statistics that no gradient step
wrote and no release records. Refitting them on kept data at bit-identical weights moves 47 of %
221 released checkpoints past the spread their own release's seeds show, several inside a %
method whose average does not move: what moves is the checkpoint's property, not its method's. %
What does the moving is not the removed data surviving in the state: exchanging kept records
for removed ones inside a fixed fitting pool moves a published cell by almost nothing, while
how far a checkpoint's shipped state has drifted from any refit does track it. The consequence
for a published decision is real but narrow: twelve verdicts cross, four clear a measured
recalibration budget, two clear it on every replicate, and a population we trained and sited
near its own criterion yields none. A release should therefore name the fitting convention beside the
number, on the batch-normalized vision models where this channel exists.
\end{abstract}


\section{Introduction}

A machine unlearning procedure claims to remove the influence of a subset of the training
data from an already-trained
model \citep{cao2015towards,bourtoule2021machine,guo2019certified}. Whether it succeeded is
settled by an audit. It compares the unlearned checkpoint against a reference retrained
from scratch without the removed data. Three published quantities decide it: accuracy on the
removed data, on the data that was kept, and on a held-out test
set \citep{golatkar2020eternal,triantafillou2024are}. A pass/fail criterion is computed
from the three, and all of it is read as a property of the released checkpoint. But a
deployed network is not only its trainable tensors. Every batch-normalization layer also
ships a running mean and a running variance \citep{ioffe2015batch}, fitted from data, used
at every inference, and written by no gradient step.

The two arms of that comparison do not come by this state the same way. The difference is
built into the comparison rather than left there by an oversight. The retrained reference
fits its statistics on retained data \emph{by construction}, because retained data is all it
ever saw. An unlearned checkpoint ships whatever its own procedure's buffer-update mode left
behind. For some procedures that is the base model's statistics untouched and for others is
a running average accumulated while the weights moved. No release we measured records which
index set, transform or accumulation rule produced the statistics it ships, so a reader
cannot recover the convention from the release. So we put both arms on one named convention and
report what moves. Take a checkpoint, change no trainable weight, refit only that state from
data the release kept, and recompute the numbers and the verdict the release published for it.
Three findings follow, and the third is the one this paper is named for. The published numbers
move. They move per checkpoint rather than per method, so a method-level verdict does not
describe the artifact an auditor holds. And what moves them is a statistic gone stale against
weights that moved, rather than the removed data surviving inside it.

Two questions are separable here and we keep them apart throughout. Whether the two arms of an
audit stand in the same relation to this state is a question only a release shipping the
retrained reference can answer, and one of ours does. Whether a published cell is sensitive to
the convention at all is a question any release can answer, and it is what the census below
measures.

\begin{figure}[t]
\centering
\begin{tikzpicture}[
  font=\scriptsize, x=1mm, y=1mm,
  bx/.style={rounded corners=2pt, draw=black!55, thin, inner sep=2.5pt,
             align=center, text width=#1, anchor=north west},
  hd/.style={font=\scriptsize\bfseries, text=black!65, anchor=south west},
  ar/.style={-{Latex[length=4pt]}, draw=black!60, semithick},
]
\node[hd] at (0,2) {WHAT AN AUDIT COMPARES};
\node[bx=27mm, fill=black!4] (rt) at (0,0)
  {\textbf{retrained reference}\\[1pt] $\theta_{\mathrm{ref}}$};
\node[bx=27mm, fill=black!4] (rp) at (30,0)
  {$\phi$ fitted on \emph{retained data}\\[1pt] by construction};
\node[bx=22mm, fill=black!3] (rc) at (63,0) {published\\cell};
\node[bx=17mm, fill=black!3] (rv) at (90,0) {pass\,/\,fail};
\draw[ar] (rt.east) -- (rp.west);
\draw[ar] (rp.east) -- (rc.west);
\draw[ar] (rc.east) -- (rv.west);

\node[bx=27mm, fill=orange!8] (ut) at (0,-13)
  {\textbf{unlearned checkpoint}\\[1pt] $\theta$};
\node[bx=27mm, fill=orange!12] (up) at (30,-13)
  {$\phi$ = \emph{whatever its run left}\\[1pt] provenance unrecorded};
\node[bx=22mm, fill=black!3] (uc) at (63,-13) {published\\cell};
\node[bx=17mm, fill=black!3] (uv) at (90,-13) {pass\,/\,fail};
\draw[ar] (ut.east) -- (up.west);
\draw[ar] (up.east) -- (uc.west);
\draw[ar] (uc.east) -- (uv.west);

\node[hd] at (0,-27) {WHAT WE DO};
\node[bx=27mm, fill=black!4] (tt) at (0,-29)
  {$\theta$ \emph{byte-identical}\\[1pt] on both arms};
\node[bx=27mm, fill=blue!8] (tp) at (30,-29)
  {$\hat\phi$ refit on \emph{retained data}\\[1pt] one convention, named};
\node[bx=22mm, fill=black!3] (tc) at (63,-29) {recomputed\\cell};
\node[bx=17mm, fill=black!3] (tv) at (90,-29) {pass\,/\,fail};
\draw[ar] (tt.east) -- (tp.west);
\draw[ar] (tp.east) -- (tc.west);
\draw[ar] (tc.east) -- (tv.west);
\draw[ar, dashed, draw=black!45] (up.south) -- (tp.north);
\end{tikzpicture}
\caption{\textbf{An unlearning audit compares two arms whose deployed state was fitted on
different data, and only one of them was fitted on retained data by construction.} No release
records which retained subset, transform or accumulation rule produced the state it ships, so
the term is invisible to an audit reading outputs and parameters. Our test refits $\phi$ on
retained data at each checkpoint's own byte-identical tensors (Section~\ref{sec:indexed}).}
\label{fig:teaser}
\end{figure}
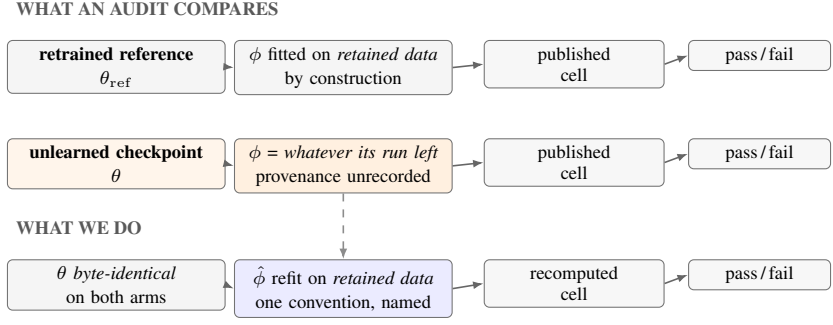

A fair objection is that an auditor should read the artifact as shipped, so the state a run
left is part of what that run produced. We agree, and attribute none of the shipped state to
the procedure that wrote it. The asymmetry is on the other side: the reference's state is
fitted on retained data whatever one thinks of the unlearned one. No removal procedure we are
aware of targets, standardizes, reports or audits the variable that separates them
(Figure~\ref{fig:teaser}). The channel also fixes the reach. Convolutional vision models ship
these running statistics. LayerNorm and RMSNorm backbones, including the language models most
current unlearning work targets, ship none, so nothing measured here carries to them.

\paragraph{Conventions.}
Three fitted states have to stay distinct. A released checkpoint \emph{ships} one; an
omission retrain \emph{accumulates} one online over retained data while its weights still
move; our test \emph{refits} one after the fact at frozen weights. A checkpoint is
\emph{anchored} when it reproduces its own published numbers with its shipped state
untouched, and one that fails this never enters an aggregate. A \emph{family}, or \emph{arm}
where the contrast with the reference is the point, is the set of checkpoints a release ships
for one removal procedure. A \emph{cell} is one entry of a release's published
table, and each family has published accuracies on the removed subset ($d_f$), the
retained subset ($d_r$) and the test set ($d_t$). A \emph{shift} is refitted minus shipped,
and \emph{displaced} means shifted past a stated margin. A \emph{crossing} is a checkpoint whose
release's own pass/fail decision changes under the refit. The \emph{budget} is how far a
cell moves when nothing was removed at all (Appendix~\ref{app:glossary}).

The margin throughout is $\pm 1.2$ percentage points (pp). It is our choice rather than a %
criterion either release publishes, fixed before the first wave at the primary release's own %
median published seed spread, and Section~\ref{sec:census} reports what happens when it moves.
Once the design below is in place, Table~\ref{tab:headline} collects every question this paper
asks beside the answer it returns.

How far a number moves is a property of the checkpoint rather than of the procedure that wrote
it, so the deliverable is a per-checkpoint test rather than a family verdict. At three variant
clusters a family mean is unresolvable at any margin, while one checkpoint's shift is
measurable to a tenth of a point. We therefore rebuild each checkpoint ten times over
independently re-drawn kept subsets and decide it on its own ten alone. Most
checkpoints sit near zero, the median moved one
at $2.247$\pp, and a family whose own interval is equivalent to zero still holds eight %
checkpoints past the margin. %

What moves a number is a shipped statistic gone stale against weights that moved, not the
removed data surviving inside it, and we measure that rather than assert it. Holding the
fitting pool at one size and exchanging kept records for removed ones moves a
checkpoint's published forget accuracy by $+0.015$\pp. It moves $+0.057$ with every removed %
record in the pool, against that $1.2$\pp\ margin. What does move with a checkpoint's shift is %
how far the state it ships has drifted from any retain-only refit (Section~\ref{sec:mech}).

\paragraph{Contributions.}
Independently of and concurrently with this work, \citet{bnillusion2026} show that this refit can
reverse a removal verdict on models they train themselves; whether the tables the field has
already published survive it is what we answer, on the checkpoints the field released.
\begin{itemize}
\item \textbf{How much a published number moves is a property of the individual checkpoint, not
  of the method it belongs to.} We run the test over 263 released checkpoints, each reproducing %
  its own published values first. We decide 221 of them on ten re-drawn replicates by an exact
  sign test, under false-discovery control that assumes nothing about how they covary.
  47 are displaced, and eight of those sit inside a method whose own interval is equivalent to %
  zero. A method-level verdict therefore bounds an average, not the artifact a reader holds. %
\item \textbf{What moves it is a stale statistic rather than surviving removed data, and we
  measure the alternative rather than argue it away.} Holding weights, estimator and pool size
  fixed and exchanging kept records for removed ones moves a published forget accuracy $+0.057$ %
  even with every removed record in the pool. Neither provenance arm moves the deployed %
  state out of a by-construction null band. Distance from a checkpoint's own refit does move,
  clears that band on all twenty, and is what separates the most displaced checkpoints from %
  the least. Nor is that distance standing in for weight movement: at equal weight distance, a %
  procedure that mixes removed data into its buffers still moves its frozen-buffer twin
  $+33.80$\pp\ further than one that does not. That is a curve reading over every rung. The one %
  distance-matched pair whose frozen arm also stays admissible cannot resolve an effect of any
  size, and we report that reading as unresolved.
\item \textbf{The convention reaches published pass/fail decisions, and we bound how far it
  reaches rather than leaning on it.} Our refit is a second defensible convention and not a
  correction, so a crossing shows that a verdict is sensitive to a choice nobody records.
  Twelve crossings survive re-decision on the release's own published cell, four exceed a %
  recalibration budget measured on the same instrument, and two do both while holding on all %
  ten re-keyed refits. No ordered method pair reverses on forget accuracy; three of 32 reverse %
  on test. %
\item \textbf{The field does not record the convention, and no cheap screen tells an auditor
  whether it matters for the checkpoint in hand.} Of 52 unlearning papers whose text we hold, %
  42 pass an inclusion rule fixed in advance, 19 run a batch-normalized backbone against a %
  retrained reference, and four of those 19 mention the state at all. None of three %
  checkpoint-readable predictors transfers across methods. Where ground truth exists we check
  our own convention against it, on the ten omission-retrained references a release ships. The
  refit reproduces their shipped state to a median $0.0019$ of a channel's own activation %
  spread, against a $0.10$ bar fixed in advance. %
\item \textbf{The account carries to populations we built ourselves, and the verdict half of
  it does not.} On a fourth population, a different dataset and architecture sited near its %
  own published criterion so that a crossing was reachable, 20 of 45 checkpoints displace past %
  that population's own $1.868$\pp\ margin. None crosses, the nearest missing by %
  $0.204$\pp. Static quantization, a fitted deployed state that every architecture has %
  including the ones with no running statistics at all, moves none of 118 checkpoints. %
\end{itemize}

\section{Deployed state and the audits that omit it}
\label{sec:related}

\paragraph{Removal is certified from outputs and parameters.}
Unlearning evaluations compare an unlearned artifact against an omission-retrained reference
and read accuracy over the removed and retained splits, often with a membership-inference
score beside it \citep{golatkar2020eternal,shokri2017membership,triantafillou2024are}. A large
literature varies that readout, the attack behind it, or the enforcement game an audit sits
in, and Appendix~\ref{app:related} places this paper against each line. Their variable is the
metric, the attack or the enforcement game, where ours is a state channel
all of those metrics read. \citet{li2026rethinking} come closest
in kind, calibrating the audit metric itself against a retrained policy, where we hold the
metric fixed and ask what the reference's own fitted state contributes to it. The two compose:
a calibrated metric read off arms with mismatched state inherits the same term. Closest in
spirit is \citet{thudi2022necessity}, arguing an unlearning claim cannot be settled from a
released weight vector at all. \citet{thudi2022unrolling} taxonomise what makes an approximate
removal verifiable, and \citet{yu2025impossibility} show retrain equivalence unattainable once
training is multi-stage. Their variable is the training history behind the weights, ours a
fitted state those weights do not contain. The two failures of the reference compose
rather than compete.

That gap is measurable, so we measured it. Of 52 unlearning papers whose text we hold, 42 %
pass an inclusion rule fixed in advance. Of those, 19 run batch-normalization-bearing %
convolutional backbones against a retrained reference. Four of the 19 mention the state %
anywhere in their text --- three once the concurrent work that is itself about it is set %
aside \citep{purge2026}. This is a statement about what a paper \emph{documents}, not what its %
authors did (Appendix~\ref{app:protocol}).

\paragraph{Normalization statistics are a variable others re-estimate for other reasons.}
Recomputing population statistics at frozen weights is a standard operation and not one we
propose: weight averaging, data-free quantization and pruning all run
it \citep{izmailov2018averaging,nagel2019data}. \citet{wu2021rethinking} establish that the estimator used for those statistics
changes accuracy. Their exact-statistics recomputation is the rebuild we adopt rather than
dispute; their variable is the estimator, whose converged same-distribution prediction is a
small offset shared across models. Test-time adaptation recomputes the same statistics to
absorb covariate shift \citep{schneider2020improving,nado2020evaluating,wang2020tent,li2017revisiting}.
That changes the data those statistics see. We hold the distribution fixed and change only
which kept subset the state was fitted on. Neither the operation nor the staleness it exposes
is new, and one concurrent paper has now named the confound outright. What is new is what the
operation is used to measure: that the movement it produces is indexed by the individual
checkpoint rather than by the method, and that it is a stale statistic rather than surviving
removed data. Both are decided on third-party artifacts against the tables their own authors
published.

\paragraph{Concurrent work on this channel, and the dated record of this one.}
An unlearned model can be perturbed until it remembers. Quantization does it, retain-set
fine-tuning does it, and in concurrent work so does recalibrating deployed normalization state
on self-trained models \citep{zhang2025catastrophic,durableun2026,siddiqui2025dormant,purge2026}.
The diagnostic is older than the papers describing it. It ships in \citet{purge2026}'s released
code as \texttt{experiments/bn\_recalibration.py}, added in the commit of 24 June 2026 and
indexed in that repository's README as a BN-recalibration attack diagnostic. That paper's arXiv
v1 of 2 June 2026 does not mention it. The instrument and protocol here were developed
independently of it: bars pre-registered 29 August 2026, census rows minted from 30 August 2026,
and the second deployed-state channel pre-registered 6 September 2026
(Appendix~\ref{app:timeline}).
\citet{bnmemorization2026}, posted May 2026, report that normalization statistics amplify
memorization during training. \citet{bnillusion2026}, posted 8 September 2026, formalize the same
weight-preserving recalibration and attribute the gap to the running statistics rather than the
weights, on nine methods they train themselves, where it reverses headline forget accuracy by up
to 78 points. The two records agree where they overlap: our normalization-free stratum returns
zero, our membership readout on the primary release does not move, and recoveries of their order
appear on checkpoints that fail their own release's criterion. That is why every count below is
read on the admitted stratum.
Deciding whether the field's published tables survive that operation needs the released
artifacts themselves, each held to the numbers its own authors printed and re-decided on its
own release's criterion. The direction of the finding differs with the object as well. Forget-class
accuracy stays at $0.0000$ wherever we move utility, so nothing here recovers removed %
content (Appendix~\ref{app:related}). %

\section{A reconstruction test for released artifacts}
\label{sec:method}

\paragraph{The operation.}
The test asks one thing of a released checkpoint: how much of the numbers it published
survives a retain-only refit of its deployed state. Let $\theta$ be a checkpoint's trainable
parameter tensors and $\phi$ its deployed normalization state, the per-channel running mean
and variance each batch-normalization layer applies at inference. We hold $\theta$
byte-identical and replace $\phi$ with $\hat{\phi}$, computed by a single forward pass over a
sample of the checkpoint's own retained data. First, $\hat\phi$ uses exact statistics rather
than an exponential moving average.
It is therefore a function of the index set and $\theta$ alone, which we verify by refitting under
permuted order and a different batch size. Second, the images must pass through the
\emph{producer's training transform} and not its evaluation transform, and
Appendix~\ref{app:instrument} reports what using the wrong one costs. Accuracies are written
as fractions throughout, matching the releases' own tables, and shifts in percentage points, so
a criterion bound of $0.05$ is five points wide. Writing $m(\theta,\phi)$ for a published %
quantity, the \emph{shift} we report is $m(\theta,\hat\phi) - m(\theta,\phi)$, and a
checkpoint is \emph{displaced} when that shift clears a stated margin on every replicate.

\paragraph{The rebuild is a convention, and we measure what that costs.}
Our refit is not a re-enactment of the reference arm's own operation. An omission retrain
accumulates its statistics online while the weights still move, whereas we compute exact
statistics once at fixed final weights. What the two share is the data the state is fitted on,
the variable an audit leaves uncontrolled. One artifact class measures that gap. Against the
ten omission-retrained references the second release ships, our refit of the same checkpoint
sits at a median channelwise distance of $0.0019$ against a $0.10$ bar fixed in advance. The %
unlearned families sit $1.8$ to $47.1$ times further away (Appendix~\ref{app:controls}). %

\paragraph{Populations.}
We audit the two releases we could find that publish a per-checkpoint table an outsider can
reproduce, and only one ships the omission-retrained reference the field compares against.
The primary population is the MU-Bench release of CIFAR-100 ResNet-50 unlearning
checkpoints \citep{cheng2024mubench,he2016deep}. Of its 180, 72 are admitted by the
benchmark's own criterion $|d_f - d_t| \le 0.05$. They span three families over three %
procedure variants, five removal ratios and five %
seeds \citep{chundawat2023can,graves2021amnesiac}. A fourth,
SalUn \citep{fan2023salun}, is admitted by none of its own. The same release ships CIFAR-10
groups on ResNet-50 and on a basic-block ResNet-34. The second is the independently authored
Unlearning Comparator release of CIFAR-10 ResNet-18 class-deletion
checkpoints \citep{lee2026unlearning}, five arms over ten classes including selective synaptic
dampening \citep{foster2024fast}, and it carries the reference. Table~\ref{tab:design} maps
every stratum to the claim it carries and the count it carries it at. Two totals recur: 263
checkpoints anchored to their own published cells and 221 of them carrying the replicate %
census. %

\paragraph{Anchoring.}
No artifact enters an aggregate until we reproduce its own published numbers as shipped. On
the primary population 71 of 72 admitted artifacts reproduce their published $d_f$, $d_r$ and %
$d_t$ within $0.02$, and on the second 42 of 50 do, all 8 exclusions being SalUn. An absolute %
anchor binds the whole primary population (Appendix~\ref{app:census}). SalUn fails to anchor %
wherever it appears, %
on 54 of its 55 CIFAR-10 cases on \emph{forget} accuracy alone while its other two cells %
reproduce normally. That is a difference in which items are scored rather than a loading %
failure. No SalUn checkpoint enters any count below (Appendix~\ref{app:census}).

\paragraph{The equivalence margin, and what a family design can resolve.}
We test for equivalence and not only for difference. Verdicts are two one-sided tests at the
90\% level against the $\pm 1.2$\pp\ margin, computed per %
family \citep{schuirmann1987comparison,lakens2017equivalence}. Because treating a
release's grid of variants, ratios and seeds as independent observations would understate how
variable a family mean is, every primary-population interval below models that structure at
the widest of three estimators (Appendix~\ref{app:clustering}). That makes a third outcome
possible, and naming it is what keeps a null honest: an interval spanning the margin resolves
neither verdict, and we call that family \textsc{unresolved}. Which outcome a family can reach
is a property of its cluster geometry, which we measure by planting a known constant into its
own centred shifts.

\paragraph{One checkpoint, end to end.}
Take the MU-Bench bad-teaching checkpoint at $2\%$ removal: as shipped it reproduces its %
published cells and the criterion reads $|d_f - d_t| = 0.0414$, inside the $0.05$ bound. Refit %
its state on retained data at unchanged tensors and the same criterion reads %
$0.0573$ (Table~\ref{tab:worked} follows three more). %

\section{What the rebuild moves, and what it leaves alone}

\begin{table}[t]
\caption{\textbf{Some published unlearning numbers do not survive a retain-only refit of the
deployed state, and what predicts which ones is how stale that state has become.} Every
question this paper asks, the population it is answered on, and the answer, with the bound
that rides with it. The sections below give the design behind each row.}
\label{tab:headline}
\centering\footnotesize\setlength{\tabcolsep}{4pt}
\input{tables/headline}
\end{table}

\subsection{How far a published number moves is indexed by the checkpoint}
\label{sec:indexed}

The second population carries the difference evidence (Table~\ref{tab:second}). It is the one
release that ships the omission-retrained reference the field itself uses. The reference and
CF-$k$ both come back
\emph{equivalent} to zero at $+0.014$ and $+0.053$\pp. AdvNegGrad does not, at $+1.916$\pp\ %
with an interval of $[+1.49, +2.34]$, and paired within deleted class over three re-keyed %
replicates it is $+1.879$\pp\ against the reference on the same class, same-signed on all ten %
classes the release ships. SSD's much larger $+11.382$\pp\ is set by three checkpoints that recover tens of points of %
retain-class accuracy while failing their release's own criterion, so nothing we claim rests %
on them. %

The primary population is a grid rather than a set of independent checkpoints, so its family
intervals are the clustered ones (Appendix~\ref{app:clustering}). Two of its three families are
\textsc{equivalent} to zero on every metric they publish, under nulls that are powered rather
than silent. The third moves, at point estimates of $+2.469$, $+2.234$ and $+1.859$\pp, but its %
family mean is not resolvable on this release. Over three procedure-variant clusters the %
half-width reaches $\pm 2.591$\pp, at which geometry a planted true zero also comes back %
unresolved, so we withdraw that verdict rather than replace it with a null. What the release %
does support is a statement per checkpoint. One checkpoint of that family shifts its own
forget cell $+3.827$\pp\ at a replicate spread of $0.061$. %

\subsection{A census: family equivalence does not imply per-checkpoint stability}
\label{sec:census}

Two quantities set the scale for any count here, the spread of one checkpoint across
replicates and the margin a reader of the release's own table would already tolerate. Both
were fixed before any count was read. For each anchored checkpoint we run the rebuild ten
times, re-keying
\emph{both} nuisance draws, the retained subset the state is fitted on and the augmentation
stream it sees. We then ask whether that checkpoint's own displacement clears the same
$\pm 1.2$\pp\ margin the family verdicts use. Nothing is pooled, and displaced always means %
positively displaced, the test being one-sided. That covers 221 of the 263 anchored %
checkpoints across two releases, two datasets and three convolutional architectures, and %
coverage is complete rather than selected: all 221 anchor at every one of the ten draws. %

\paragraph{The decision rule, and what it assumes.}
Because the two nuisance draws are re-keyed independently, a checkpoint's ten displacements
are independent. Under the boundary null that its median displacement sits exactly at the
margin the signs of the ten deviations are fair coins. Requiring all ten to land beyond the
margin on the same side is then an exact one-sided sign test at level $(1/2)^{10} = 0.00098$, %
with no normality, no resampler and no fitted constant. Its price is that this is also the %
smallest $p$-value it can return, so the census is a floor rather than an estimate.
Multiplicity is controlled across all 221 at once by Benjamini-Yekutieli under arbitrary
dependence, so what we claim is false-discovery control (Appendix~\ref{app:controls}). The ten
draws re-key the refit's own nuisance, so what is certified is within a checkpoint; training-run
variability enters only as the yardstick the margin is set from.

\paragraph{What the census finds.}
Of the 221 checkpoints, 47 are individually displaced on forget accuracy, 39 on retain and %
40 on test, every one certified at $q = 0.05$ (Table~\ref{tab:percensus}). Inside those counts %
is the dissociation no family-level verdict can express. The family that is \textsc{equivalent}
to zero on all three metrics under the primary clustered estimator still contains 8 of its %
45 checkpoints displaced past the same margin. That is arithmetic rather than a surprise: a %
family mean of $+0.830$\pp\ against a $1.2$\pp\ margin, with real across-checkpoint spread, %
must put some of them past it. The audit therefore has to be run on the checkpoint in hand
rather than on its family (Appendix~\ref{app:clustering}). These are not marginal cases: the
displaced checkpoints sit at a median $2.247$\pp, against a generic recalibration cost of %
$0.736$\pp, and accuracy headroom is excluded by measurement %
(Appendix~\ref{app:negatives}). %
Neither free parameter carries the result. Moving the margin
to $1.0$ or $1.5$\pp\ takes the forget count to 56 or 31, monotonically on all three metrics. %
Quadrupling the retained pool moves no checkpoint across it (Appendix~\ref{app:sens}). %

Neither the accuracy readout nor the metric family carries it. Run on the other quantity this
release publishes about forgetting, the census displaces 21 of 72 checkpoints past that %
release's own spread (Appendix~\ref{app:zrf}). Against the one release shipping a real %
omission-retrained oracle per deleted class, one family of four moves $+0.0167$ of the %
base-to-oracle separation back toward the un-unlearned base on 10 of 10 classes. Ten classes %
sharing one base training is all that licenses (Appendix~\ref{app:drift}).

\subsection{What indexes it is a stale statistic, not the removed data}
\label{sec:mech}

The tempting reading is that a shipped statistic still carries the data the removal was meant
to delete, and we measured that arm rather than arguing it away. Hold the weights, the estimator, the image transform and the size of the fitting pool fixed.
Then build three pools that share a retain core and differ only in where the remaining $k$
records come from. One draws them from a disjoint retain reserve, one from the removed set at
the share a deployment would draw, one from the removed set in full. The first two differ in the provenance of $k$ exchangeable
records and in nothing else, so the contrast is exact rather than controlled
(Appendix~\ref{app:expose}).
Provenance moves the published number by almost nothing (Table~\ref{tab:headline}). At the
deployment share the pooled forget cell moves $+0.015$\pp, and $+0.057$ with every removed %
record in the pool, against a census margin of $1.2$. No checkpoint reaches $0.5$\pp\ on %
either contrast and the largest per-checkpoint mean is $+0.271$. This is a measurement and not %
a failure to look, since the same maximal-provenance arm on the quantization channel of
Section~\ref{sec:ptq} does move three checkpoints past their margin.

A null on a published number has two readings: the state changed and the audit could not see it,
or it did not change. Measured in the channel-standardized units of
Section~\ref{sec:method}, both provenance contrasts sit inside the band those references
fixed, at 0 and 1 of 20 checkpoints above it. The distance between a checkpoint's shipped
state and any retain-only refit clears it on all twenty and is $5.43$ times larger at %
the median. That distance is also the term that moves with the displacement: across the two %
strata, the ten most displaced checkpoints sit much further from their own refit than the ten
least. Their provenance distance is identical to the fifth decimal. So what varies across checkpoints is how stale the
state they ship has become. The arms had already suggested it: CF-$k$ ships the base model's
statistics with the removed class still in them and comes back equivalent to zero
(Table~\ref{tab:second}). What separates the arms is whether a procedure updated the
buffers while it moved the weights, not whose data those buffers saw. Both provenance figures
are conservative upper bounds (Appendix~\ref{app:expose}).

One rival survives that measurement: that the state only stands in for how far the weights
moved, since weights that moved further leave whatever shipped with them staler. We put that
to an intervention rather than a correlation, on the self-trained population of
Section~\ref{sec:gen}. Freezing only the buffer update yields a twin bitwise weight-identical
to its sibling, 90 pairs of 90 sharing a trainable-weight hash and differing in their buffer %
hash. The gap inside a pair is therefore the deployed state and nothing else. A six-rung %
optimization ladder per procedure then meets an exposed and an unexposed one at equal weight
distance. There the exposed arm's twin gap stays $+33.80$\pp\ above the unexposed %
arm's own distance curve, over five seed clusters at $[21.28, 46.32]$. The one pair matched on %
distance alone whose frozen twin also stays admissible cannot settle it. Both arms sit at the %
forget metric's own quantum against a $12.93$\pp\ detection floor, so admissibility and %
resolvability do not overlap on this producer (Appendix~\ref{app:ladder}). What the ladder %
isolates is the trajectory those buffers were accumulated along rather than whose data entered
the pool, since the two arms' fitting streams share most of their images and a normalization
statistic never reads a label.

\subsection{Published verdicts cross the threshold in both directions}
\label{sec:verdicts}

The rebuild changes the decision each release publishes about its own checkpoints, in both
directions and on both releases (Table~\ref{tab:verdicts}). MU-Bench admits a
checkpoint when $|d_f - d_t| \le 0.05$, a gap between two of its own cells with no retrained %
reference in it. What moves those verdicts is therefore the staleness term of %
Section~\ref{sec:mech} and not the asymmetry between arms. Under the rebuild 8 of its 71
natively admitted CIFAR-100 checkpoints fall out of that criterion, 4 of the 42 %
anchor-passing excluded ones fall in, and its CIFAR-10 groups add 2 more. On the second %
release, whose criterion is a different one, two of four natively failing checkpoints pass %
with nothing traded away.

We bound this rather than lean on it. Two screens are reported as a conjunction rather than
composed: whether a crossing survives being re-decided on the release's own published cell
instead of our recomputed one, and whether it clears a recalibration budget
(Appendix~\ref{app:crossings}). That budget is measured on this instrument rather than
borrowed: no-removal checkpoints give $0.698$\pp, which the spread across ten %
independently trained oracles moves to $0.736$. The crossings the budget does not cover sit %
inside it and are not separable from generic recalibration, so this is an existence result on
named checkpoints at the field's own operating point, not a rate. Re-decided on each of the ten
refits the census already carries, both hold on all ten under either rule %
(Table~\ref{tab:xstab}). Two boundaries run the other way. No ordered method pair reverses on %
forget accuracy. No checkpoint-readable screen substitutes for running the test: of three predictors fixed before
their inputs existed, two fail within every family. The third, weight distance from a
checkpoint's own base, is significantly positive in 4 of 9 families and in none of the %
primary population's three (Appendix~\ref{app:negatives}). %

\subsection{How far the account reaches, and where it stops}
\label{sec:gen}

No third release we could find publishes a per-checkpoint table an outsider can reproduce, so
we built two populations instead, both self-trained and pooled with neither release. The first
is SVHN and VGG-16-BN through a public implementation at its own
defaults \citep{netzer2011reading,simonyan2015very,fan2023salun}. There the displacement
reappears on the one arm that mixes removed and retained data, at $+0.473$\pp\ against that %
population's own $0.323$\pp\ margin, an order below the released magnitudes %
(Appendix~\ref{app:gen}). %

The second population answers what a small effect cannot. It is CIFAR-100 on the same
architecture, sited near the published $|d_f - d_t| \le 0.05$ criterion by a rule fixed before %
any base existed. A population far inside that bound cannot produce a crossing at any
displacement. The magnitude arrives at released scale: of the 45 checkpoints its producer's
default policy ships, 20 are displaced past that population's own $1.868$\pp\ margin on ten %
re-keyed refits, reaching $+13.26$\pp\ on one. The verdict does not, and its absence has a %
measured cause rather than a shortage of power. None of the 45 crosses, against a budget of %
$0.085$\pp\ from this population's own bases. The rebuild lifts the forget cell further than %
the test cell, so it widens the gap on 33 of the 45 while 30 sit above the bound where only a %
narrowing converts. The nearest checkpoint missed by $0.204$\pp\ (Appendix~\ref{app:reach}). The %
claim that a verdict crosses rests on the two released populations, and this bounds it. %

Which families move is predicted rather than only described. We read each release's own
unlearning loop for whether it updates buffers over batches holding removed data, blind to any
displacement and with both tests fixed beforehand. That reading separates exposed from
unexposed families by $1.865$\pp, and replicates prospectively at the SVHN population's %
zero-exposure cells (Appendix~\ref{app:bnmode}). That population also carries the paper's %
interventional arm: a twin freezing only the buffer update, bitwise weight-identical to its
sibling, differs from it by $-9.50$ to $-75.81$ points of test accuracy. %
Section~\ref{sec:mech} reads that arm at matched weight movement. %

\paragraph{A second deployed state, on backbones that carry no running statistics.}
\label{sec:ptq}
Normalization statistics are one instance of a class, and unlike them the activation
scales a deployment fits when it quantizes exist on any architecture. Run unchanged over 118
anchored checkpoints, 47 of them swin-base transformers, that census returns zero on every %
metric, under a familywise correction the accuracy census cannot reach %
(Appendix~\ref{app:ptq}). %

\section{Limitations}

\textbf{No family-level difference is established on either population, and the family that
moves cannot be resolved at all.} The second population's AdvNegGrad interval is
\textsc{different} against the margin, but that is displacement under our own rebuild rather
than a difference between procedures, and its ten runs share one base training. A family mean
on the primary release rests on three procedure-variant clusters, where the detection floor
returns a planted zero and a planted $+3.0$\pp\ alike. On the second the paired contrast %
tracks each checkpoint's own accuracy level at $-0.918$ (Appendix~\ref{app:clustering}). %

\textbf{The census is a floor, not an estimate, and its multiplicity control is
false-discovery rather than familywise} (Section~\ref{sec:census}). \textbf{What stays
unexplained is which procedural detail leaves one family's shipped state staler than
another's, and whether that detail is inherent to a removal algorithm or a choice its
implementation made.} The classifier of Section~\ref{sec:gen} reads an implementation
property, a training loop's buffer-update mode, and nothing separates that from the algorithm
it implements. The mixture identity predicting a rebuilt statistic from the removed fraction
fails all three of its pre-declared bars (Appendix~\ref{app:negatives}). What the ladder of
Section~\ref{sec:mech} buys is bounded on two sides. The state's contribution is not a
function of how far the weights moved, but it attributes to the trajectory those buffers were
accumulated along rather than to whose data they saw. Its matched and admissible reading is
unresolved rather than positive.

\textbf{The forgetting-quality evidence is one family of one release:} ten classes sharing a
single base training, with the split that concentrates the effect declared after the first
draw. The shift is $+0.0167$ against an across-class spread of $0.0119$ %
(Appendix~\ref{app:drift}). %
\textbf{The verdict evidence is an existence result, not a rate,} and it stands on the two
released populations. Most MU-Bench crossings sit inside the $0.736$\pp\ our measured budget %
allows, and on a population we built and sited so a crossing was reachable, none of 45 %
checkpoints crosses. Raising it needs checkpoints sitting just below their own criterion. %

\textbf{The first instrument was wrong,} and about half of the originally measured effect was
the instrument: correcting it took the largest family shift from $+5.02$\pp\ %
down to $+2.47$ (Appendix~\ref{app:instrument}). \textbf{A displacement is therefore defined %
relative to a rebuild convention,} and no single magnitude is interpretable without naming
that convention. \textbf{The third population establishes reach, not a comparable effect
size,} resting on four admissible artifacts of one arm; the fourth closes that half at
released scale and returns a null on the verdict. \textbf{And the reach beyond normalization
is measured as a null.} The census stays convolutional and batch-normalized across three
codebases, three datasets and four architectures, none of which a layer-normalized model
joins \citep{ba2016layer,wu2018group}. The second channel covers one operating point, and
calibration temperatures, feature scalers and class priors are untouched on any pipeline.
Nothing here says how far a published cell of a language model moves.

\section{Consequences for deletion audits}

\paragraph{What follows for a release.}
Three requirements are usually collapsed into one, and separating them is what makes this
record actionable. \textbf{Preserving} the state is done: both releases serialize the buffers. \textbf{Recording} its fitting history is cheap \citep{abdullah2025scale}, but only the
producer can do it.
\textbf{Reporting} the cell
under a named rebuild convention is what the evidence supports, because the shipped number is
one side of a comparison whose two arms were fitted on different data, and quoting it alone
hides a term the reader cannot recover. Concretely, a release reporting an audit cell should
add one line: \emph{deployed state as shipped; refit on the retained split under the training
transform of \texttt{train.py}, cell moves $+2.47$\pp}. We recommend neither overwriting the %
shipped state nor standardizing one convention across releases.
Reversals of the size concurrent work reports on self-trained models appear here only outside the
admitted stratum: the three checkpoints recovering the most retain-class accuracy, up to
$+58.63$\pp, all fail their own release's criterion. That regime is one the benchmarks' own %
admissibility excludes, and on admitted released artifacts the effect is small, indexed by the
checkpoint, and not screenable.

An unlearning audit compares two models whose deployed state was fitted on data no release
records. Putting them on one convention at bit-identical weights moves published numbers per
checkpoint, and not because the removed data survives. What that licenses is reproducibility,
not deletion safety.

\clearpage

\subsection*{Ethics statement}
This work audits publicly released model checkpoints and their published evaluation
tables. It involves no human subjects, no new data collection and no personal data.
It reports that one released arm's published forget cell is not reproducible under the
conventions we could recover. Other released cells, while reproducible as
shipped, are not invariant to a benign recomputation of the state those artifacts ship.
Both are statements about measurement validity, not about the intent or competence of
their authors. We name arms rather than individuals, and report the
anchor failures as counts alongside everything else. The failure mode we disclose is one an auditor needs in order to read a deletion claim correctly. It does not supply a capability for recovering removed content: forget-set accuracy is unchanged
wherever we move utility. Deletion audits are used to substantiate legal erasure
obligations \citep{foschvillaronga2017humans,ginart2019making}. A verdict that moves
under a retain-derived recalibration of the state it ships is a risk to the party
relying on it.

\subsection*{Reproducibility statement}
The measured objects are public third-party releases, which is what makes this
result checkable independently of our code. Section~\ref{sec:method} states the rebuild, the anchoring rule and the equivalence
margin. Appendix~\ref{app:recipe} gives the rebuild verbatim, including the image
transform, the accumulation rule, the retained-pool size, the seeds and the exact
repository identifiers of every checkpoint used. Appendix~\ref{app:census} gives the population census and per-family
anchor pass rates. Appendix~\ref{app:clustering} gives every family interval under every
estimator we ran, together with the planted-signal detection floor.
Appendix~\ref{app:instrument} gives the instrument crossing and the
superseded values, and Appendix~\ref{app:crossings} every verdict crossing by name.
Appendix~\ref{app:ptq} gives the second deployed state: the quantization scheme, the
calibration sizes, the operating point and the bar that failed.
Every table and figure in this paper is generated by a script that reads the stored
per-run result artifacts: \verb|tables/make_tables*.py| and \verb|figs/make_figures.py|,
each naming the result files it reads at the top of every emitting function. Those
scripts and artifacts are included in the supplementary code archive.
What the archive lets a reader do with a single released checkpoint is the whole chain this
paper runs on it. Fetch it by the repository identifier we record, and reproduce its published
cells as shipped to confirm that it anchors. Refit its deployed state on the retained split
under the named convention, and recompute those same cells and the pass/fail verdict its
release computes from them. Then regenerate every table and figure here from the stored
artifacts. The cost is measured rather than estimated. One refit-and-score pass over 15
checkpoints took 3368 seconds on a single A100 MIG slice, so reproducing a complete ten-draw %
census for one checkpoint costs under 40 GPU-minutes. %

\subsubsection*{Acknowledgments}
This work was supported by the CIFAR AI Chairs program, Amii, and NSERC.

\bibliography{references}
\bibliographystyle{iclr2027_conference}

\appendix
\input{appendix}

\end{document}

%% file: math_commands.tex
\usepackage{amsmath,amsfonts,bm}

\def\eqref#1{equation~\ref{#1}}

\def\1{\bm{1}}

\DeclareMathAlphabet{\mathsfit}{\encodingdefault}{\sfdefault}{m}{sl}
\SetMathAlphabet{\mathsfit}{bold}{\encodingdefault}{\sfdefault}{bx}{n}



%% file: tables/headline.tex
\begin{tabular}{@{}p{0.395\textwidth}rp{0.36\textwidth}@{}}\toprule
question & $n$ & what the test returns \\\midrule
do the published numbers reproduce as shipped, before any test? & 263 & these 263 do, within $0.02$; the rest are excluded \\
does forget accuracy move past $\pm 1.2$\pp\ on ten re-drawn refits? & 221 & 47 displaced (39 retain, 40 test) \\
does a second forgetting measure move, on three refits? & 72 & 21 displaced \\
does a published pass/fail verdict cross? & 263 & 14 candidates $\to$ 12 on the published cell $\to$ 4 above the measured budget $\to$ 2 on both \\
do \emph{removed} records in the fitting pool move a cell? & 20 & $+0.015$\pp\ pooled, 0 past $0.5$\pp \\
has the shipped state drifted from any retain-only refit? & 20 & $0.00538$ against a $0.0019$ null band, on all 20 \\
does a second deployed state move a cell (quantization scales)? & 118 & 0 displaced, 47 of them transformers \\
does it reproduce on a fourth population, sited near its own criterion? & 45 & 20 displaced past that population's own $1.868$\pp\ margin; 0 verdicts cross, the nearest missing by $0.204$\pp \\
is the state's effect a proxy for how far the weights moved? & 90 & no: $+33.80$\pp\ residual at matched movement, $[21.28, 46.32]$ \\
does the rebuild reorder two removal procedures? & 32 & 0 reverse on forget, 3 on test \\
\bottomrule\end{tabular}

%% file: appendix.tex
\section{Population census and anchoring}
\label{app:census}

Both populations are third-party public releases, and every artifact is anchored to
its own published table before it enters any aggregate. The anchoring rule is a
tolerance of $0.02$ on each of the three published quantities. Table~\ref{tab:census} %
gives the census by stratum and family.

The MU-Bench release carries 180 ResNet-50 CIFAR-100 artifacts across four
procedures. Of these, 72 are admitted by the benchmark's own criterion
$|d_f - d_t| \le 0.05$, 73 fail it while publishing usable values, and 35 ship no %
usable result file and therefore have no anchor at all. SalUn is admitted 0 of 45 by
that criterion and is excluded from the admitted stratum entirely; when we measure
the excluded stratum in Section~\ref{sec:verdicts}, the same arm accounts for 30 of
the 31 anchor failures there. The eight anchor failures on the second population are
also SalUn, eight of that arm's ten artifacts. The same arm therefore fails on two
independent releases, but it fails them differently, and the difference is the whole
of what we can claim. On MU-Bench the failure is confined to a single published cell:
of the 55 failing artifacts, 54 fail on forget accuracy alone, while their retain and
test cells reproduce to $10^{-4}$ and $4\times10^{-4}$ respectively --- the same order
as every other family, which fails 0 of 150 on all three %
cells. On the second release the failure is broader, reaching both cells that release
publishes for this arm, 6 of 10 on forget and 6 of 10 on retain. We therefore do not
claim these weights fail to reproduce their published
table. What we claim for MU-Bench is narrower and is all the evidence carries: this
arm's forget cell is not recoverable without the arm's own forget-set definition,
which no release in either population documents. That is a fact about the arm and not
about our loader: the other families anchor at 71 of 72 on the admitted stratum, 42 of 43 on
the excluded stratum, and 40 of 40 on the second release.

\begin{table}[h]
\caption{\textbf{The one arm whose state was fitted on retained data by construction is the
one that returns to zero.} Retain-class accuracy shift under the rebuild, Unlearning
Comparator, ten deleted classes per arm; \textsc{e} is an \emph{equivalent} verdict and
\textsc{d} a \emph{different} one. Only two of SalUn's ten checkpoints anchor, and SSD's mean
is set by three that fail their release's criterion (Appendix~\ref{app:census}).}
\label{tab:second}
\centering\footnotesize\setlength{\tabcolsep}{3pt}
\input{tables/second}
\end{table}

\begin{table}[h]
\caption{\textbf{Which stratum carries which claim, and at what count.} \emph{Anchored}
counts checkpoints reproducing their own published cells as shipped; \emph{in census} counts
those decided on ten re-keyed replicates. The second release enters through the paired
contrast rather than the census, which is why $263$ anchored checkpoints yield a census %
of $221$. The natively excluded stratum is measured separately, for the inward crossings
only, and the quantized stratum is a different deployed state on an overlapping population;
neither is counted in either total.}
\label{tab:design}
\centering\footnotesize\setlength{\tabcolsep}{4pt}
\input{tables/design}
\end{table}

\begin{table}[h]
\caption{Population census, the per-family detail behind Table~\ref{tab:design} rather than a
second set of counts. \emph{Artifacts measured} counts what entered the
pipeline; the next column counts those reproducing their own published $d_f$, $d_r$
and $d_t$ within $0.02$ as shipped. Summing a stratum's families here %
reproduces that stratum's row above.}
\label{tab:census}
\centering\scriptsize
\input{tables/census}
\end{table}

Two absolute anchors bind the pipelines to something outside themselves. On the
primary population the released pre-unlearning base checkpoint recomputes to a test
accuracy of $0.8328$ against its published $0.8328$; the two nearby preprocessing %
variants we tried give $0.8217$ and $0.8131$, so the anchor is discriminating and not %
a coincidence. %
On the second population the anchor error is $0.0002$ and the rebuild is additionally %
verified to be a function of the index set alone: refitting under permuted order and
under a different batch size changes the recomputed accuracy by $0.0$ percentage %
points. %

\begin{table}[h]
\caption{\textbf{Family-level equivalence does not imply per-checkpoint stability: neggrad
is \textsc{equivalent} to zero on all three metrics under the primary clustered estimator
and still carries 8 displaced checkpoints.} Checkpoints whose displacement is positive past
the $\pm1.2$\pp\ margin, by the exact one-sided sign test on each checkpoint's own re-keyed %
draws. The last column is that family's own clustered verdict on forget accuracy, so the
dissociation is readable in one place, and it is blank for the CIFAR-10 groups because that
release publishes no per-family criterion for them. Rows marked \emph{8 rep.} are what the
eight-replicate wave records; the ten-replicate aggregate stores pooled counts only, so
eight is the finest by-family resolution available. The two total rows are the pooled
counts, and only the ten-replicate one is certified by Benjamini-Yekutieli at $q=0.05$. The %
ten-replicate displaced set is a subset of the eight-replicate one by construction, so the
extra draws cost one checkpoint on forget accuracy and none on retain or test.}
\label{tab:percensus}
\centering\footnotesize\setlength{\tabcolsep}{4pt}
\input{tables/percensus}
\end{table}

\paragraph{The two numbers we chose rather than measured.}
\label{app:sens}
The census has two free parameters. One is the margin a displacement is scored against; the
other is how many retained examples the state is refitted on. Neither is estimated from the
data, so we swept both on the frozen draws rather than argue for the values we picked.
Table~\ref{tab:sens} reports the sweeps. Moving the margin from $1.2$ to $1.0$\pp\ raises %
the
forget count from 47 to 56 and moving it to $1.5$ lowers it to 31, monotonically and on all %
three metrics. At $2.0$\pp\ the count is 23 and none of it is certified, which is %
multiplicity arithmetic rather than checkpoints ceasing to move: every rejected $p$-value
equals the sign test's floor, so the Benjamini-Yekutieli step needs a tied set of at least 26
and 23 is short of it. The fitting pool matters less. Across 4000, 8000 and 16\,000 %
examples the per-checkpoint displacement agrees with the value the paper carries to a median
$0.150$ and $0.100$\pp, every one of the 71 checkpoints stays inside the margin, and the %
family mean is flat at $+1.181$, $+1.191$ and $+1.190$\pp. One pre-registered clause failed %
and we report it as failed: a single-draw threshold count on test accuracy moves 28, 22 and %
25 across the three pools, outside the band we declared. That failure is the reason the
carried census uses ten re-keyed draws and an exact test rather than one draw.

\begin{table}[h]
\caption{The census against its two free parameters, on the frozen ten-draw displacements.
Panel (a) counts displaced checkpoints at four margins, given as the count at the sign
test's floor and the count Benjamini-Yekutieli certifies at $q = 0.05$; the two differ only %
at $2.0$\pp, where the tied set falls below the threshold the procedure needs. Panel (b) %
varies the number of retained examples the state is refitted on, one draw per pool size, and
compares each against the 8000-example pool the paper uses.}
\label{tab:sens}
\centering\footnotesize
\input{tables/sens}
\end{table}

\section{The state rebuild, stated exactly}
\label{app:recipe}

The contribution rests on one operation, so we give it here rather than a
hyperparameter table. Let $\theta$ be the trainable tensors of a released
checkpoint, loaded with \verb|strict=True| from the release's own
\verb|model.safetensors|, and let the batch-normalization layers of the model be
$\ell_1 \dots \ell_L$ in topological order. The rebuild replaces each layer's running
mean and variance and touches nothing else.

{\footnotesize
\begin{verbatim}
for target in batchnorm_layers(model):   # topological order
    model.eval()                         # every BN stays in EVAL mode
    n, s, ss = 0, None, None
    for b in loader(pool, sorted(idx), bs=256, shuffle=False):
        x = input_to(target, b)          # forward, early-exit at target
        dims = [0] + spatial_dims(x)
        s  = add(s,  x.sum(dim=dims, dtype=float64))
        ss = add(ss, (x.double()**2).sum(dim=dims))
        n += x.numel() // x.shape[1]
    target.running_mean = s / n
    target.running_var  = (ss - s*s/n) / (n - 1)    # unbiased
    target.num_batches_tracked = 0
\end{verbatim}}

Three properties of this procedure decide whether the contrast means anything,
and each was verified rather than assumed.

\paragraph{It is a function of the index set and $\theta$ alone.}
Layers are fitted sequentially, so layer $\ell_{k}$ sees inputs that already reflect
$\ell_{1..k-1}$'s finished statistics. The obvious alternative is to run the network in
training mode with momentum \verb|None| and let the framework accumulate, and it is
\emph{not} order-independent. That path averages per-batch means and per-batch
unbiased variances, so the result moves with the batch partition. In training mode
each layer's input distribution depends on batch composition as well. We checked
the exact procedure against permuted order and a different batch size and confirmed
that the recomputed accuracy is unchanged.

\paragraph{The pool carries the producer's own training transform.}
MU-Bench accumulated its shipped statistics over the transform
\begin{center}\small\verb|RandomResizedCrop(224) + RandomHorizontalFlip + ToTensor + Normalize|\end{center}
read from its training script rather than from its preprocessor configuration file,
which disagrees with the script and loses. Scoring instead uses the same release's
evaluation transform,
\begin{center}\small\verb|Resize(224) + CenterCrop(224) + ToTensor + Normalize|\end{center}
which is the one that reproduces the published base accuracy exactly. Using the
evaluation transform for the \emph{rebuild} costs more than two percentage points of
instrument, as Appendix~\ref{app:instrument} shows.

\paragraph{Every scored item is held out of the fitting pool.}
The retained pool is 8000 training images drawn without replacement from the retained
indices; the retained-accuracy probe is a disjoint 10000 images drawn from the same
pool of candidates, with the intersection asserted empty in code. An arm whose scored
items sat inside the refit pool would buy an advantage its counterpart cannot have,
and that advantage would land inside the contrast.

\paragraph{Determinism and versions.}
The pool is materialised once per augmentation seed and shared across every artifact
in a shard, so all artifacts in a comparison see byte-identical inputs. Augmentation
seeds are $\{0, 1, 2\}$ for the family-level waves and the second metric family, and every
reported family-level shift is the mean over them; the across-seed spread on the base
checkpoint is $0.052$ percentage points on test accuracy, which is the instrument noise a %
single seed inherits. The per-artifact census does not use that triple. It re-keys both %
nuisance draws ten times per checkpoint, the retained subset as well as the augmentation
stream, and decides each artifact on its own ten. %
Rebuild batch size is 256 and does not affect the result. After each rebuild the code
asserts that the SHA-256 of the concatenated trainable tensors is unchanged and that
the SHA-256 of the buffers has changed; a violation of either aborts the run.
Checkpoints are the \verb|jialicheng/*| ResNet-50 CIFAR-100 repositories of MU-Bench
and the ResNet-18 CIFAR-10 class-deletion checkpoints of Unlearning Comparator, in
both cases at the versions publicly available in 2026. Every artifact used is named
in the supplementary artifact index.

\section{The controls, collected}
\label{app:controls}

Section~\ref{sec:census} states each rival explanation and the measurement that
excludes it. Table~\ref{tab:controls} collects them in one place, including the two
known-answer negative controls, one per population, that check the scoring pipeline
does not report structure where there is none.

\paragraph{The per-artifact test's detection floor.}
The floor is measured at the geometry the census actually runs at. A planted $1.5$\pp\ %
effect is recovered on 134 of the 150 CIFAR-10 checkpoints and $2.0$\pp\ on all 150, while %
plants at or below the threshold recover none. One defect in our own earlier controls is
worth flagging. A one-sided lower-bound rule cannot reject at a planted mean of exactly
zero whatever the data, so those planted-zero arms verify the implementation and nothing
more, which is why we measure the size at the threshold instead. The census wave itself
reproduces, returning all 70 comparable checkpoints to within $0.000$\pp\ when re-run from %
scratch.

\begin{table}[h]
\caption{\textbf{Every rival that would explain the shift away is measured; three are
smaller than the effect or point the wrong way, and the fourth is unresolved.} Controls,
the reading each one addresses, and what it measured. The native-accuracy rival is
separated by the sign of the contrast but not by its interval once the release's own
clustering is modelled, and is reported as \textsc{unresolved} rather than excluded. The first four use the primary population; the last
two are known-answer negative controls, one per population.}
\label{tab:controls}
\centering\scriptsize
\input{tables/controls}
\end{table}

\paragraph{Why ten replicates and not eight.}
The exact sign test can only reject at $(1/2)^{k}$ on $k$ draws, so the multiplicity
procedure available to it is a function of the replicate count. At eight draws the floor
$p$ of $0.0039$ sits above the Benjamini-Yekutieli threshold, which certifies zero %
checkpoints and leaves only Benjamini-Hochberg, whose positive-dependence condition a weak %
measured correlation does not establish; Benjamini-Hochberg at that count would have
carried 48. At ten draws the floor of $0.0009766$ clears Benjamini-Yekutieli, so the %
carried counts hold under arbitrary dependence with no assumption about how the checkpoints %
of one release covary. It does not clear Bonferroni, and thirteen replicates would be
needed for that, so no familywise claim is made anywhere in this paper.

\paragraph{The census count is not checkpoint-to-checkpoint scatter.}
Independently trained checkpoints differ from one another before any rebuild, so one
reading of the census is that it measures that scatter. We price the rival rather than
argue it: the scatter would need a standard deviation of $0.746$\pp\ to put even the %
smallest displaced artifact past the census margin at two standard deviations, which is
$6.2\times$ the upper bound the three MU-Bench base checkpoints give and $10.3\times$ the %
ten independently trained oracles on the second population measure. This is a breakdown %
bound rather than a pre-registered test, and the MU-Bench term is an upper bound rather
than a measurement, which is why it is reported here and not as a headline.

\paragraph{The rebuild convention, against the one state class with ground truth.}
The second release ships ten omission-retrained references whose deployed state was itself
accumulated over retained data while the weights moved. We compare each shipped state
against our post-hoc exact refit of the same checkpoint, channel by channel, in units of
that channel's own activation standard deviation, and pre-registered a threshold of $0.10$ %
on the median. The references come in at $0.0019$. The unlearned arms sit further away by %
factors of $1.8$, $17.5$, $19.3$ and $47.1$, the reference band disjoint from the nearest %
of them, and that ordering matches the ordering of the metric shifts across all five arms. %
The displacement is therefore visible in the state tensors and not only in an accuracy
cell. Two scope notes: this is one draw per checkpoint, so the spread is across checkpoints
rather than across draws; and one arm's factor is a state measurement rather than an
anchored one, only 2 of its 10 checkpoints reproducing their own published cells under the %
forget-set convention we could recover. %

\paragraph{The calibration we withdrew.}
We tried to establish the Student-$t$ rule's finite-sample size by resampling each
artifact's centred replicates and fitting a critical value that brings the rejection rate
to 5 percent. Fed Gaussian residuals, for which the rule is exact, that machinery returns a
size of $0.0643$ where the answer is $0.050$ and fits $2.499$ where the answer is $2.132$, %
larger even than the $2.449$ it fits on real residuals. A bootstrap-$t$ on five centred %
residuals is itself anti-conservative, so we had measured the resampler rather than the
rule, and a split-sample check cannot catch that because both halves share the artefact.
The calibration is withdrawn, no carried count rests on the $t$ rule, and the census is
decided by the exact sign test of Section~\ref{sec:census} instead.

\section{The second metric family}
\label{app:zrf}

The primary release publishes a zero-retrain-forgetting score beside its accuracy cells. It
compares an unlearned artifact's output distribution on the removed data against a randomly
initialized model's, so it reads the same deployed state through a different statistic. We
ran the census on it unchanged: three replicates per checkpoint, both nuisance draws
re-keyed, decided against a margin $M = 0.00277$ taken from the release's own published %
spread, with a wider sensitivity margin of $0.00438$ reported beside it. %

Three bars were fixed before any of those numbers were read. The plumbing bar asks that
this wave's recomputed \emph{accuracy} shifts reproduce the frozen accuracy wave, and they
do, on 71 of 71 checkpoints at a worst difference of $0.0$\pp. The anchor bar asks that the %
new metric reproduce its own published values, and 70 of 71 do, at a median absolute error %
of $0.0036$. The wave covers all 72 artifacts the benchmark admits, rather than the 71 of %
the accuracy census, because it anchors on its own metric instead of inheriting the accuracy %
anchor: 71 of the 72 publish a zero-retrain-forgetting value at all. %
The no-removal bar asks that a checkpoint carrying no removal provenance stay %
inside the margin, and it moves $-0.0007$. Table~\ref{tab:zrf} gives the result. The %
measurement is precise against its own margin, the median replicate standard deviation
being $0.00031$, and the moving family's mean shift of $+0.0115$ is about four times the %
margin. %

Attribution stays withheld here exactly as it does on accuracy. The bar we pre-declared was
that an artifact's displacement not track its native accuracy level; the correlation is
$-0.30$ pooled but reaches $-0.56$ and $-0.65$ inside two of the three families. So this %
corroborates the descriptive family pattern on an independent readout and licenses no
causal claim. It is also the same batch-normalization state, so it widens the readout and
not the channel reach.

\begin{table}[h]
\caption{\textbf{An independent forgetting measure separates the same families more sharply
than accuracy does.} Zero-retrain-forgetting shift under the rebuild, MU-Bench CIFAR-100,
three replicates re-keying both nuisance draws. $M$ is the release's own margin and
$M_{\mathrm{s}}$ the wider sensitivity margin. The last row is the no-removal base
checkpoint, this metric's own negative control.}
\label{tab:zrf}
\centering\footnotesize
\input{tables/zrf}
\end{table}

\section{The second deployed state: quantization scales}
\label{app:ptq}

\paragraph{What is fitted, and how.}
Static post-training quantization maps activations to 8-bit integers through one scale per
tensor, fitted by passing calibration data through the network and reading an order statistic
off each site's activation range. We attach the quantizer by forward pre-hook, so trainable
tensors and normalization buffers are provably untouched and the only thing that differs
between arms is which records the calibration set holds. The convolutional group carries 52 %
quantized sites per checkpoint and the swin-base group 149. Both use the absmax criterion at %
512 calibration records, an operating point fixed on a checkpoint with no removal provenance %
before any unlearned artifact was quantized. Quantizing at that point moves forget accuracy
by $-1.65$\pp\ on the convolutional group and $-0.21$\pp\ on the transformer group, %
so the quantized model is a usable one rather than a degenerate one. %

\paragraph{Two contrasts, and why one of them is exact.}
The displacement contrast is the one the accuracy census uses: each published cell recomputed
under a retain-only calibration set against the cell the checkpoint produces as shipped,
decided against the release's own $\pm 1.2$\pp\ margin over thirteen re-keyed draws. The %
identification contrast is the stronger one. Both of its arms draw the same retain core and
differ only in whether $k = \mathrm{round}(n_{\mathrm{cal}} n_{\mathrm{del}} /
N_{\mathrm{train}})$ of their records come from the removed set or from a matched retain
reserve. Under the null of no dependence on removal provenance the two arms are exchangeable,
so the sign of the paired difference is a fair coin by construction rather than by
assumption. Thirteen draws on one side is then a $p$-value of $0.000244$, and the threshold %
that controls the familywise rate over the 118 checkpoints is $0.000424$. The exact test %
therefore has the resolution to certify familywise here, which at ten draws it does not.

\paragraph{Controls, including one bar that failed.}
The pre-unlearning base checkpoint of each group carries no removal provenance and sits at
$0.048$ and $0.042$\pp\ of absolute mean displacement. A nuisance contrast that redraws the %
same number of records without ever touching a removed one flags no checkpoint either, where
chance alone would be expected to flag fewer than one of the 118. The detection floors in
Table~\ref{tab:ptq} are planted into each checkpoint's own per-draw values rather than into
synthetic noise, so they price this population's real spread. One bar did fail. The plumbing
bar tying this wave's unquantized pass to the frozen normalization wave asks that the two
agree on the checkpoints they share, and over those 70 the worst difference is $0.0333$\pp, %
past the bar we fixed in advance. On that checkpoint's removed split it is a single example %
changing side, and the cause is batch-size-dependent kernel selection between the two waves.
Every displacement reported here is a within-wave contrast whose two arms share that
selection, so we report the bar as failed rather than round it to a pass.

\paragraph{Why the channel is quiet, predicted before it was measured.}
The mechanism was declared in advance. Absmax is an extreme order statistic, so one record
can set a scale and its influence on that scale decays like $O(1)$ in the calibration size,
while percentile and mse are averaging criteria whose influence decays like $O(1/n)$. Going
from 128 to 2048 calibration records, the median per-moved-site log ratio falls from $0.031$ %
to $0.015$ for absmax on the convolutional group and from $0.032$ to $0.021$ on the %
transformer group, against $0.018$ to $0.004$ and $0.014$ to $0.003$ for percentile. Over the %
same range absmax moves 4 sites and then 3 on the convolutional group, and 14 and then 16 on %
the transformer group, while percentile moves 26 and then 40, and 78 and then 135. So absmax %
moves few scales by amounts that barely decay and percentile moves many by amounts that decay
fast, and neither route carries enough to reach an audit cell. The accuracy leg of that
prediction passes on the convolutional group and is unevaluable on the transformer one, where
the median displacement at the largest calibration size is exactly zero and the declared
ratio is undefined. We report it as unevaluable rather than as a pass.

\begin{table}[h]
\caption{\textbf{The same census on a second deployed state returns zero on both
architectures, including the one that ships no normalization statistics at all.} Static
8-bit activation scales refitted on a retain-only calibration set, MU-Bench CIFAR-100,
thirteen re-keyed draws per checkpoint. The consistent-sign rows use the exchangeable
contrast at a margin of zero; the nuisance row is the same rule applied to a contrast that
moves no removed record. Detection floors are the planted displacement recovered on at least
90\% of a group's checkpoints. The last two rows calibrate entirely on removed records, which
upper-bounds the channel rather than describing a deployment.}
\label{tab:ptq}
\centering\footnotesize
\input{tables/ptq}
\end{table}

\section{A third population, built from scratch}
\label{app:gen}

Both audited releases are convolutional networks on CIFAR, so the census cannot say on
its own whether the displacement is a property of that setting. No third release we
could find publishes a per-checkpoint table an outsider can reproduce, so we built a
population instead. Everything below is self-trained. It is reported as such and pooled
with neither release in any count of this paper.

\paragraph{How it was built.}
The dataset is SVHN and the architecture is VGG-16-BN, a thirteen-site plain
convolutional stack with no residual connections \citep{netzer2011reading,simonyan2015very}. The
unlearning runs go through a public implementation at a fixed commit, at its own
defaults, with one disclosed one-line repair without which the released head does not
import \citep{fan2023salun}. We train five base checkpoints under five seeds, and for each seed run
an omission retrain, a retain-only fine-tune, gradient ascent on the removed data and
masked relabelling. Each method cell is run twice: once at the producer's default,
which updates the normalization buffers while unlearning, and once with only that
update suppressed. Every design decision was written down before any artifact existed.
The margin is set the way the primary release's own margin was, as the median
across-seed spread of the population's own native metric, which here is
$0.323$\pp\ on forget accuracy. %

\begin{table}[h]
\caption{\textbf{On a self-trained third population the displacement reappears on the one
arm that mixes removed and retained data, and the controls return to zero.} SVHN,
VGG-16-BN, forty artifacts at ten re-keyed draws each. \emph{Admissible} counts the
artifacts passing both the gap criterion and a utility floor referenced to that seed's own
omission retrain; the \emph{admissible} shift column is the mean over those, and is blank
where a cell has none. \emph{Displaced} is the exact sign test at that population's own
$0.323$\pp\ margin, certified by Benjamini-Yekutieli. The \emph{frozen} rows carry weights %
bitwise identical to their default sibling and differ in the shipped buffers alone; they
fail the utility floor, so they show the state changes behaviour rather than describing a
deployable artifact.}
\label{tab:gen}
\centering\footnotesize\setlength{\tabcolsep}{3.5pt}
\input{tables/gen}
\end{table}

\paragraph{What it returns.}
Table~\ref{tab:gen} gives every cell. On the four admissible artifacts of the masked
relabelling arm the retain-only rebuild moves the forget cell $+0.473$\pp\ on average, %
with per-artifact means of $0.240$, $0.280$, $0.431$ and $0.942$, and 2 of the 4 clear the %
margin under the exact test. The three controls measured on the same instrument in the same
run return to zero: retain-only fine-tuning $-0.006$\pp\ over five artifacts, the no-removal %
base $-0.010$ and the omission retrain $+0.024$. Two limits ride with this. The magnitudes %
are an order below the released populations' displaced checkpoints and are read against a
margin roughly a quarter as wide, so this establishes reach and not a comparable effect
size. And the retain and test margins here are degenerate under the pre-registered rule, at
$0.005$ and $0.044$\pp, because self-trained cells reproduce across seeds far more tightly %
than a released benchmark's do, so no count is read on those two metrics.

\begin{table}[h]
\caption{The third population's own bookkeeping: the external anchor its base checkpoints
pass, the margins its rule produces, and the blocking invariants each artifact was checked
against. Eighteen of the forty artifacts return a single forget value across all ten draws,
which the metric's own quantization on a 4500-item forget split explains.}
\label{tab:genfacts}
\centering\footnotesize
\input{tables/genfacts}
\end{table}

\paragraph{What the criterion admits, and what the frozen twin shows.}
Table~\ref{tab:genfacts} carries a result about the criterion itself. All forty artifacts
pass $|d_f - d_t| \le 0.05$, and 19 of them sit far below their own retrain reference: a %
model at chance satisfies a gap test trivially. The sharpest case is not merely a weak
model, since gradient ascent at the producer's default diverges to entirely non-finite
weights on four of five seeds, more than 15 million values each, and those checkpoints
still satisfy the criterion. Every count in this appendix is therefore reported twice, once
under the criterion alone and once under a utility floor.

Suppressing only the buffer update leaves the twin bitwise weight-identical to its default
sibling, verified at a relative weight distance ratio of exactly $1.0$ on all arms with %
finite weights. The two therefore differ in the shipped normalization buffers and in nothing
else whatever, and freezing changes test accuracy by $-9.50$ to $-75.81$ points; a %
retain-only
rebuild then recovers $+11.119$\pp\ of forget accuracy on the fine-tuning arm against %
$-0.006$ for its default twin. That is the only interventional evidence in this paper that %
deployed state carries behaviour the released weights do not determine, and its scope is
narrower than it first reads. The utility cost of freezing is not specific to unlearning:
retain-only fine-tuning passes no removed example through the network and still moves test
accuracy by $-9.50$\pp, and the relabelling arm's larger $-75.81$\pp\ comes at a relative %
$L_2$ of $0.713$ against $0.127$. Exposure and weight movement are %
confounded across the two usable arms, so what the twin establishes is that shipping buffers
which no longer match moved weights is costly, not that the cost comes from the removed data.

\paragraph{The prediction that could not be run.}
Two predictions written before the wave failed, and we report them as failures. The
100\%-exposure cell is unavailable rather than refuted: gradient ascent at its producer's
default is inert on one seed and collapses the model to $24.5$\% test accuracy on the other %
four, which is the same wall two other public implementations produced. And the frozen twin
is unevaluable inside the admissible window, because freezing the update while the weights
move ships a stale state that costs 8 to 10 points of test accuracy, a different defect from
a contaminated one.

\section{A fourth population, sited where a verdict could cross}
\label{app:reach}

The third population answers half of what the two releases leave open. Its magnitudes are an
order below theirs, and its checkpoints sit tens of points inside their own criterion, so no
displacement it could show would ever reach a verdict. Whether a crossing is a property of
those two releases or of the phenomenon needs a population whose checkpoints sit near a
criterion to begin with, and no release we found supplies one. So we built it. Everything
below is self-trained, reported as such, and pooled with neither release in any count.

\paragraph{How it was built, and how it was sited.}
The dataset is CIFAR-100 and the architecture is again VGG-16-BN, unlearned through the same
public implementation at the same commit and at its own defaults \citep{fan2023salun}. Five
seeds each carry their own base and their own omission retrain, for 100 artifacts, each
rebuilt on ten draws that re-key the retained subset and the training augmentation
independently. The augmentation draw is live here, where the producer's SVHN recipe left it
degenerate. Siting came first and was written as code before any base existed: a ladder of
optimization strengths per procedure, then a rule picking the rungs whose native
$|d_f - d_t|$ gap sits closest to the published $5$\pp\ criterion. The nearest rung the %
population reaches sits $2.00$\pp\ from that criterion, against tens of points for the %
untuned ladder. %

\paragraph{What the anchors and controls return.}
The five bases land between $72.00$ and $72.42$ percent test accuracy, inside a band we fixed %
by reading a published table before any base was scored. The census instrument is the third
population's, bit-identical to it: over the 40 artifacts they share, every native, refit and %
shift cell agrees to $0.000$\pp\ with matching hashes throughout. Across 990 artifact-draws no %
artifact has a moved weight tensor and none repeats a refit buffer. Against the producer's own %
evaluation block over 100 artifacts the worst disagreement is $0.022$\pp\ on forget accuracy %
and $0.010$ on test, with one retain probe failing at $1.025$. The bar excludes that one %
artifact rather than rounding it to a pass, and we report the bar as partly failed. The
crossing detector does depend on which artifact each rebuild belongs to: permuting that join
returns 150 crossing artifact-draws against 50 under the true one. %

\begin{table}[h]
\caption{\textbf{On a population sited where a crossing was reachable, the displacement
arrives at released scale and the crossing does not.} Panel A decides each of the 45
checkpoints the producer's default policy ships on its own ten re-keyed refits, by the exact
sign test the released populations use, under Benjamini--Yekutieli at $q = 0.05$. Panel B is
what a crossing would have taken: the rebuild moves most of these checkpoints away from their
bound rather than toward it.}
\label{tab:reach}
\centering\footnotesize\setlength{\tabcolsep}{4pt}
\input{tables/reach}
\end{table}

\paragraph{What it returns} is Table~\ref{tab:reach}. The magnitude reproduces at released
scale, at 20 displaced checkpoints of 45 against that population's own margin and a largest %
per-checkpoint displacement of $+13.26$\pp, where the third population's admissible %
displacements ran $0.240$ to $0.942$. The verdict effect does not. None of the 45 crosses its %
criterion, and the reason is directional rather than a shortage of power: the rebuild lifts
the forget cell further than the test cell, so on 33 of the 45 it widens the gap that the %
criterion bounds, while 30 of the 45 sit above the bound, where only a narrowing would %
convert. The nearest missed by $0.204$\pp. Five crossings do exist in the same wave, all of %
them on frozen-buffer artifacts, which is our own intervention rather than anything this
producer ships; they are reported here and enter no count in this paper. %

\paragraph{Two corrections to our own analysis.}
Addendum 4 of the pre-registration records both, dated after the first aggregate ran and
before any number of it was carried into this paper. The first implementation of the crossing
count omitted the pre-registered restriction to the producer's default policy, and so reported
the five frozen-buffer crossings as though they were shipped ones. The second computed the
smallest convertible shift without the sign of the shift this effect actually produces, and
quoted a floor from an artifact the effect's own direction cannot reach.

\section{Matching weight movement between an exposed and an unexposed procedure}
\label{app:ladder}

One rival to the staleness account of Section~\ref{sec:mech} survives every measurement above:
that the deployed state stands in for how far the weights moved. On the released census that
rival came back mixed (Appendix~\ref{app:negatives}), and the frozen twin of
Appendix~\ref{app:gen} could not settle it, because the two arms usable there differ in weight
distance by a factor of five. This wave removes that difference rather than controlling for it.

\paragraph{Design.}
On the third population's own bases, each of three procedures runs along a six-rung ladder of
optimization strength, and each rung runs twice: once at the producer's default, which updates
the buffers while unlearning, and once with only that update suppressed. That is 180 cells
across five seeds. Inside a pair the trainable weights are bitwise identical and only the
shipped buffers differ, so the gap between a pair's two members is the deployed state and
nothing else. An exposed and an unexposed procedure are then matched on the median relative
$L_2$ distance from their own base, at $|\ln$ ratio$| \le 0.2$, a rule fixed before any cell of
the wave was adjudicated.

\begin{table}[h]
\caption{\textbf{At equal weight distance the exposed procedure's deployed state still carries
more, and the one matched pair that is also admissible cannot see an effect of any size.}
Panel A is the primary reading, one row per pair matched on distance alone. Panel B is the
better-powered reading over every rung, with the unmatched operating point beneath it as the
check that an effect of this kind is visible to this instrument at all.}
\label{tab:ladder}
\centering\footnotesize\setlength{\tabcolsep}{4pt}
\input{tables/ladder}
\end{table}

\paragraph{The primary reading is unresolved, and that is the finding.}
Three pairs match on distance (Table~\ref{tab:ladder}). A pair may be read only where its
frozen arm stays admissible, and one of the three qualifies, on all five seeds. There both
arms sit at the forget metric's own quantum: the largest single arm-seed contrast is
$0.533$\pp\ against a detection floor of $12.93$\pp\ measured on this forget set. The two %
pairs whose contrasts do resolve, at $+13.04$ and $+51.37$\pp, keep their frozen arm %
admissible on 3 and 0 of 5 seeds. Admissibility and resolvability do not overlap on this %
producer, which the pre-registration's first addendum predicted before the wave ran.

\paragraph{The curve reading does resolve it.}
Using every rung rather than the three matched pairs, the exposed arm's twin gap sits
$+33.80$\pp\ above the unexposed arm's own distance curve, over five seed clusters at %
$[21.28, 46.32]$. Its most direct instance is the near-matched pair at median distances %
$0.715$ and $0.652$, whose twin gaps differ by $+51.37$\pp\ with the same sign on all five %
seeds. A confirming outcome was reachable by construction: at the unmatched operating point %
the same wave returns $+65.31$\pp\ with a spread of $10.57$ and five seeds of five agreeing in %
sign, so what leaves the primary reading unresolved is the matching and not the instrument. %

\paragraph{What it isolates, and what it does not.}
The two arms of a pair differ in the trajectory their buffers were accumulated along, not in
whose data entered the fitting pool: their input streams share most of the same images, and a
normalization statistic never reads a label. This composes with the provenance null of
Section~\ref{sec:mech} rather than competing with it. Most of these rungs' frozen arms also
fail the utility floor, so this is a statement about what the deployed state carries, not
about an artifact anyone would ship.

\section{Reading buffer exposure out of the producers' own code}
\label{app:bnmode}

A family-level pattern that is only described invites the reading that we sorted the
families after seeing which moved. So we classified them from a source that could not know:
each release's own published implementation. For every family we read whether its unlearning
loop puts the network in training mode over batches that contain removed data, which is what
makes the shipped buffers accumulate over that data. The classification was fixed, along
with both tests below, before any class was compared against any displacement.

\begin{table}[h]
\caption{\textbf{Which families move is predicted by reading the producers' own code,
blind to the measurements.} \textsc{high} marks a family whose released loop updates
normalization buffers over batches containing removed data, \textsc{zero} one that never
does, \textsc{low} one that does so only on a small fraction of its steps. \emph{From}
records whether the mode was read directly from the released path or inferred from the
method's definition where the released path does not execute. $n$ counts the anchored
checkpoints that enter the family mean, not the family's released size, which is why SalUn
contributes 2 of its ten and its \textsc{high} class rests on the thinnest base in the table.
The shift column is forget accuracy on the primary release and retain-class accuracy on the
second, each family's own published metric.}
\label{tab:bnmode}
\centering\scriptsize\setlength{\tabcolsep}{4pt}
\input{tables/bnmode}
\end{table}

Table~\ref{tab:bnmode} gives the result. The magnitude test passes: the \textsc{high}
families average $+1.899$\pp\ against $+0.034$\pp\ for the \textsc{zero} families, a %
separation of $1.865$\pp\ that exceeds the margin the whole paper is scored against. The %
ordering test does not, at an exact $p$ of $0.0556$; that is the smallest value the test can %
return with three and four families, so it is a design floor rather than a negative result,
and we report it at the number it returned. Three of the classifications were inferred
rather than read because the released path does not execute as published, which is itself
worth recording for a reader who plans to re-run these releases. %

This is a prediction about families, not about artifacts. It does not say which checkpoint
inside a \textsc{high} family moves, and Appendix~\ref{app:negatives} shows no
checkpoint-readable quantity supplies that. Its independent test is prospective:
Appendix~\ref{app:gen} builds a population the classifier had never seen, and its
zero-exposure cells return $-0.006$ and $+0.024$\pp, which is what the account predicts. %

\section{Does the literature name the deployed-state convention?}
\label{app:protocol}

The main text asserts that auditing practice specifies controls over outputs and parameters and
never names non-parameter deployed state. This appendix measures that assertion on a sample rather
than leaving it as one.

\paragraph{Sample.} Every unlearning paper whose PDF we hold: the union of a corpus query for recent
unlearning evaluation work at ICLR, ICML and NeurIPS 2025--26 and the priors this paper positions
against. That is 52 unique documents, one duplicate having been dropped, and all 52 extract to %
text. Inclusion test, fixed before the widened run and unchanged from the pilot: the text contains %
``unlearn'' at least fifteen times. 42 papers pass it and 10 do not. It is a defined population %
rather than an exhaustive census of the field, and every statement we make from it carries that
rule.

\paragraph{Detector.} Each PDF is converted to text, then flattened --- hyphenation removed and all
whitespace stripped --- so that \texttt{batch-\textbackslash{}nnormalization}, ``batch norm'',
\texttt{BatchNorm} and \texttt{Batch\textbackslash{}nNorm} collapse to one string before matching.
We then search, case-insensitively, for \texttt{batchnorm}, \texttt{batchnormalization},
\texttt{runningmean}, \texttt{runningvar}, \texttt{runningstatistic},
\texttt{populationstatistic}, \texttt{bnlayer} and \texttt{bnstatistic}. The flattening exists
because a false zero here would manufacture the finding.

\paragraph{Two controls, one in each direction.} On the same detector, the concurrent work that
\emph{is} about recalibrating deployed normalization state returns 13 hits, on three of those %
terms \citep{purge2026}. A positive control is not a negative one, so we also run the detector on %
an ablated corpus with those terms removed and on a decoy term list: both return zero hits over %
all 52 papers, against 417 native hits. The detector fires when the text is there and stays silent %
when it is not, which is what makes the low counts readable. %

\paragraph{Result.} Of the 42 included papers, 19 run batch-normalization-bearing convolutional %
backbones (ResNet or VGG) and compare against a model retrained from scratch on the retained data.
Four of those 19 mention the deployed normalization state that every one of those backbones ships %
and that both arms of their comparison carry. Set aside the concurrent work that is itself about
this channel and it is 3 of 18, at 2 to 3 term hits each against that paper's 13: a %
remaining-data-free method \citep{cheng2024suppressing}, a forget-set-free %
method \citep{newatia2026forgetsetfree} and a tamper-resistance
method \citep{siddiqui2025dormant}. None of the three names it as a variable of the comparison.
The remaining 23 included papers are language models or otherwise non-convolutional, where the %
variable this paper studies does not exist.

This widened population narrows an earlier reading of ours rather than confirming it. On the
14-paper pilot the count was zero, and we retire that stronger statement: at the defined population
the answer is a small positive count, cited by name above. A paper that never names the convention
may still have applied one consistently, so this bounds what is documented and not what is
practised.

\section{Extended related work}
\label{app:related}

\paragraph{What the evaluation literature varies.}
Unlearning evaluations read accuracy over the removed and retained splits and often a
membership-inference
score \citep{graves2021amnesiac,kurmanji2023towards,jia2023model,carlini2022membership}, and
other modalities apply criteria of the same shape to generation
scores \citep{maini2024tofu,li2024wmdp,shi2024muse}. All of them are fragile to the attack
chosen \citep{hayes2024inexact,goel2022towards}. A parallel line replaces the statistic
itself \citep{seo2025revisiting,shi2025redefining,kim2026truly}, moves the intervention to
inference time \citep{chowdhury2026conformal}, enumerates the controls an audit claim
needs \citep{sommer2022athena,auditingunlearning2026,measuredontoptimize2026}, or models the
auditor-operator game a detection capability induces \citep{lin2026governing}. Each varies
the metric, the attack or the enforcement game. None of them varies the fitted state the
metric is computed through, which is the axis this paper holds a release to.

\paragraph{An unlearned artifact can be perturbed until it remembers.}
Quantizing an unlearned model can restore forgotten
content \citep{zhang2025catastrophic,durableun2026}; normalization statistics amplify
memorization during training \citep{bnmemorization2026}; an unlearned model is often only
dormant, with retain-set fine-tuning recovering the behaviour it was said to have
lost \citep{siddiqui2025dormant}; and concurrent work recalibrates deployed normalization
state on self-trained models \citep{purge2026}. Those results transform a fixed artifact
from outside or update its trainable tensors, and they measure restoration of forgotten
content. Here the checkpoint's own shipped state is the variable and the direction is
utility. Their quantization is also data-free and applied to weights, where the state
Section~\ref{sec:ptq} puts on a retain-only footing is fitted on data and applied to
activations, so the two lines do not overlap even on the channel they appear to share. The population claim is disjoint too, since we census a third party's release and
anchor every artifact to its own published table. \citet{bnillusion2026} are the nearest of
these: they formalize the same weight-preserving recalibration as a fixed-point operator,
prove that any gap it opens is attributable to the running statistics rather than to the
weights, and measure it on nine methods they train themselves, where it reverses headline
forget accuracy by up to 78 points and a normalization-free control removes it. We adopt that
operation rather than claim it. What the operator identity settles is what the gap is
attributable to. Two things it does not settle are what the published record does under the
operation, and whether the movement is the removed data surviving in the fitting pool or a
statistic gone stale against weights that moved. Both are questions about released artifacts
and their own printed tables, and both are what every count in this paper is about. Their large reversals and our small ones
are the same phenomenon read on different strata: on artifacts that fail their own release's
criterion we also recover tens of points (Appendix~\ref{app:census}), and the admitted stratum
every count here is read on is the one an auditor would actually be handed.

\paragraph{Why this state is worth auditing at all.}
Normalization state is not incidental to what a network computes. Training only the
normalization layers of an otherwise random network already reaches non-trivial
accuracy \citep{frankle2021training,santurkar2018how}, so a channel that no release records
and no audit matches is carrying part of the function being audited.

\paragraph{Changing the readout is not changing the state.}
\citet{kim2026truly} re-examine the protocols themselves and argue that logit-level readouts
overstate forgetting, recommending representation-level ones instead. That changes which
readout an audit trusts, where we hold the readout fixed and change which fitted state
accompanies bit-identical weights. Our own second metric family
(Appendix~\ref{app:zrf}) makes the two directions separable: it swaps the readout while
holding the state channel fixed, and the displacement survives.

\paragraph{Benchmark numbers are known to vary, but not like this.}
\citet{bouthillier2021accounting} bound how far seed and data resampling move a benchmark
result, and the reproducibility literature documents how often a reported number fails to
survive re-implementation \citep{pineau2021improving,raff2019step,lucic2017are}. Those are
stochastic sources that trial averaging removes; a state-estimator mismatch is systematic
and invisible to averaging. Reproducibility tooling for unlearning standardises the
pipeline rather than the deployed state a checkpoint ships \citep{dangelo2025how}. Two
recent audits adopt the discipline we follow, admitting no claim that does not exceed a
measured null on a frozen model \citep{xu2026phantom,zhu2026reading}.

\section{Terms used throughout}
\label{app:glossary}

\begin{figure}[h]
\centering
\includegraphics[width=\textwidth]{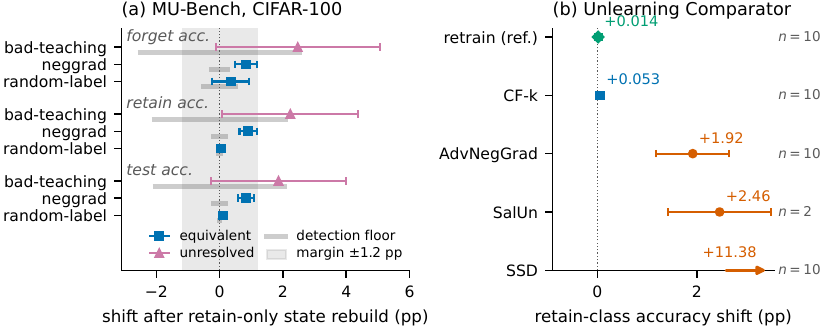}
\caption{\textbf{At identical architecture, dataset, distribution and estimator, one
family's published numbers return to within a tenth of a point while another's move past
the release's own margin.} (a) MU-Bench:
90\% two-one-sided-test intervals per family, clustered on the release's own procedure %
variants, against the $\pm 1.2$\pp\ margin. The grey bar behind each interval is that %
family's detection floor, the half-width the same estimator returns on a planted true
zero. Where the floor is wider than the margin, no verdict is available in either
direction. (b) Unlearning Comparator: retain-class accuracy shift per arm, mean $\pm$ one
standard deviation. SSD is off the scale and drawn as an arrow, its mean being set by
three artifacts that fail their release's criterion.}
\label{fig:dispersion}
\end{figure}

\begin{description}
\item[\emph{arm}, \emph{family}] the set of checkpoints a release ships for one removal
procedure. On the primary release an arm spans three procedure variants, five removal
ratios and five seeds, and we also call it a family.
\item[\emph{anchoring}] reproducing a checkpoint's own published numbers while its state
is untouched. A checkpoint that fails this never enters an aggregate.
\item[\emph{the rebuild}] replacing a checkpoint's deployed normalization statistics with
statistics fitted on retained data, at bit-identical trainable tensors.
\item[\emph{refit displacement}] the signed change the rebuild produces in a checkpoint's
own published metric. It is a measured difference between two fitted states, and no part
of it is attributed here to the removal procedure that wrote the shipped one.
\item[\textsc{different}, \textsc{equivalent}, \textsc{unresolved}] a family interval
falling strictly outside the $\pm 1.2$\pp\ margin, strictly inside it, or spanning it. %
\item[\emph{detection floor}] the half-width the clustered estimator returns when a true
zero is planted into a family's own cluster structure. A family whose floor exceeds the
margin can reach no verdict in either direction. Figure~\ref{fig:dispersion} plots both
populations' family intervals against those floors.
\end{description}

\section{Clustering, and what each family's design can resolve}
\label{app:clustering}

The primary release ships each family as a grid of three procedure variants, five
removal ratios and five seeds, so its artifacts are not independent draws. Every
family interval in the main text is estimated with that structure modelled, and
Table~\ref{tab:estimators} gives all of them under every estimator we ran: the
unclustered interval, cluster-robust CR2 variances clustered by variant, by variant
crossed with ratio and by seed, and a crossed hierarchical fit with random intercepts
for variant and ratio, read by parametric bootstrap. The main text reports the wider
of the two variant-based CR2 intervals. Verdicts hold where the design is thick and
move where it is thin. The two equivalent families keep their verdict in 29 of the 30 %
cells, the exception being neggrad retain accuracy under the hierarchical fit. The
moving family's verdict instead depends on which grouping is assumed. It is different
on all three metrics when artifacts are grouped by seed, on two of the three when they
are grouped by variant crossed with ratio, and unresolved on all three when they are
grouped by variant alone. The pre-declared rule is to report the wider of the two
variant-based intervals, which for that family is the variant grouping, so the
conservative reading is the one the main text carries. The dependence is real and sits mostly on the ratio
axis: on the moving family the intraclass correlations run $0.15$ to $0.23$ by variant %
and $0.34$ to $0.37$ by ratio, against $0.00$ to $0.03$ and $0.37$ to $0.47$ on neggrad. %

\begin{table}[h]
\caption{\textbf{Two of three procedure families are \textsc{equivalent} to zero on every
metric they publish; for the third, three procedure variants cannot resolve a family mean
in either direction.} Shift of each published quantity when deployed normalization state
is rebuilt from retained data at bit-identical trainable tensors, MU-Bench. Brackets give
90\% two-one-sided-test intervals against the $\pm 1.2$\pp\ margin, from cluster-robust %
CR2 variances over $G$ procedure-variant groups at the Student-$t$ critical value for
$G-1$ degrees of freedom. The last column is the pre-unlearning checkpoint, which has no
removal provenance by construction.}
\label{tab:primary}
\centering\footnotesize\setlength{\tabcolsep}{3pt}
\input{tables/primary}
\end{table}

\begin{table}[h]
\caption{Every family-level interval under five estimators, MU-Bench.
\textsc{d} is \emph{different}, \textsc{e} \emph{equivalent} and \textsc{u}
\emph{unresolved} against the $\pm 1.2$\pp\ margin. The main text uses the wider of %
the two variant-based cluster-robust columns.}
\label{tab:estimators}
\centering\scriptsize\setlength{\tabcolsep}{4pt}
\input{tables/estimators}
\end{table}

A widened interval that contains zero can mean the effect is absent or that the design
cannot see it, and those are different claims. To separate them we plant a known
effect into each family's own cluster structure and re-run the estimator on it. Each
family's real per-artifact shifts are centred to a true mean of exactly zero, which
preserves every variant, ratio and seed grouping and every within-cluster spread, and
a constant of $0.0$ or $3.0$\pp\ is added. Table~\ref{tab:planted} reports what comes %
back. The estimator recovers the planted level exactly in all eighteen arms. On
neggrad and random-label it also returns the right verdict at both plants, which is
what makes their nulls in Table~\ref{tab:primary} powered rather than silent. On
bad-teaching both plants come back unresolved, so at three variant clusters no effect
in the region of interest could have been resolved either way. The reason is
substantive rather than a sample-size accident: bad-teaching's shift depends strongly
on the release variant while neggrad's does not, so the family that moves is exactly
the family whose mean three clusters cannot pin.

\begin{table}[h]
\caption{Detection floor by family and metric. Each family's own per-artifact shifts
are centred to a true mean of zero, a known constant is planted, and the clustered
estimator is re-run on the real cluster structure. \emph{Recovered} is the level the
estimator returns for the planted constant.}
\label{tab:planted}
\centering\scriptsize\setlength{\tabcolsep}{4pt}
\input{tables/planted}
\end{table}

\section{The instrument correction}
\label{app:instrument}

An earlier version of this measurement rebuilt statistics on evaluation-transformed
images. It was wrong, and about half of the effect it reported was itself the
instrument. The defect announced itself on a control artifact that by construction
should show nothing and instead showed $-2.55$ percentage points. %

\begin{figure}[h]
\centering
\includegraphics[width=\textwidth]{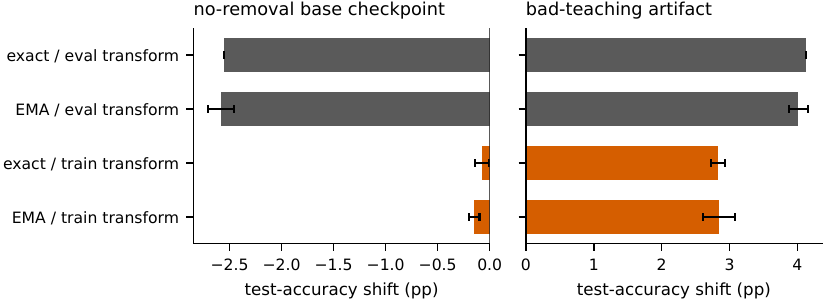}
\caption{\textbf{The offset a naive rebuild reports is the data distribution, not the
estimator's functional form.} Test-accuracy shift under all four combinations of
accumulation rule and image transform, on the released pre-unlearning checkpoint
(left) and on an artifact from the family that moves (right). Bars are means over
three augmentation seeds with one standard deviation; the exact/eval arm is
deterministic.}
\label{fig:estimator}
\end{figure}

Crossing the accumulation rule against the image transform on the pre-unlearning base
checkpoint localises the offset (Figure~\ref{fig:estimator}). Under
evaluation-transformed images the checkpoint loses $-2.550$\pp\ of test accuracy with %
exact statistics and $-2.583$\pp\ with an exponential moving average. Under the %
producer's training transform the same checkpoint sits at $-0.077$\pp\ and %
$-0.150$\pp. %
The accumulation rule is worth hundredths of a point; the transform is worth two and
a half points. The same crossing on an artifact from the moving family separates
differently, at $+4.130$\pp\ under the evaluation transform and $+2.827$\pp\ under %
the training transform. The correction therefore does not subtract a constant, and %
the family separation is not an artifact of matching only to the base checkpoint.

All 113 artifacts that were anchor-passing when the correction landed --- the 71 primary and
the 42 second-population ones, before the CIFAR-10 groups joined the census --- were
re-measured through the corrected rebuild, and every number in the main text, over all 263,
comes from it.
Table~\ref{tab:superseded} prints the superseded per-family values beside the current
ones so the size and direction of the correction are visible. It shrank the largest
family shift, moved two families from negative to positive, and moved the second
population's omission-retrained reference from $-1.44$ to $+0.01$ percentage points. %
No offset is subtracted anywhere in the current instrument, because there is no
longer an offset to subtract.

\begin{table}[h]
\caption{Superseded evaluation-transform values beside the distribution-matched
values reported in the main text, MU-Bench, per family and per published metric.}
\label{tab:superseded}
\centering\footnotesize
\input{tables/superseded}
\end{table}

\section{Every verdict crossing, by name}
\label{app:crossings}

Table~\ref{tab:verdicts} summarises the crossings by release and native status, and
Table~\ref{tab:worked} follows four named artifacts end to end, including one the test
cannot measure at all.

\begin{table}[h]
\caption{\textbf{A release's own success verdict is not invariant to putting both arms'
deployed state on the same retain-derived footing.} Crossings under the rebuild at bit-identical
trainable tensors, by release and native status. The last column gives how far each
MU-Bench crossing lands past the $0.05$ bound, over it when leaving the admitted set %
and under it when entering, and for the second release the \textsc{ra} values its
crossings move between.}
\label{tab:verdicts}
\centering\footnotesize
\input{tables/verdicts}
\end{table}

\begin{table}[h]
\caption{\textbf{What the test does to four named artifacts, including the one class of
artifact it cannot measure.} Each row rebuilds a released checkpoint's deployed
normalization state from retained data at bit-identical trainable tensors and
recomputes the quantity its release's criterion reads. The last row is the
case the test cannot reach: this arm's forget cell does not reproduce under our
forget-set convention, and since the criterion is a function of that cell the arm is
dropped before the test runs. Its retain and test cells reproduce normally.}
\label{tab:worked}
\centering\footnotesize
\input{tables/worked}
\end{table}

\begin{figure}[h]
\centering
\includegraphics[width=\textwidth]{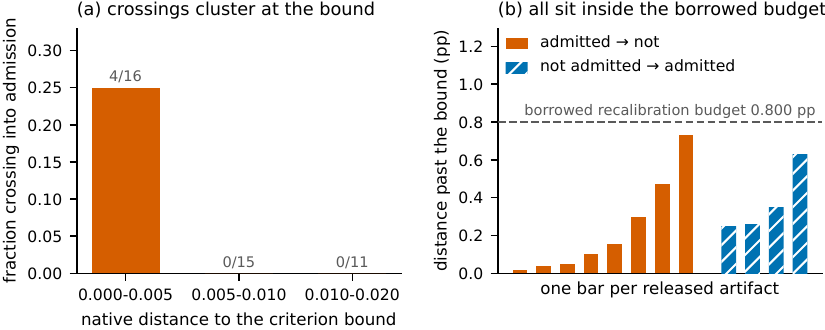}
\caption{\textbf{Verdict crossings are real and confined to a narrow band around the
criterion, and on the CIFAR-100 release none of them exceeds what a generic
recalibration could produce.} (a) Fraction of natively excluded MU-Bench artifacts
crossing into admission, by how far the artifact started from the $0.05$ bound; counts %
are crossings over stratum size. (b) How far past the bound each crossing on this
release lands, in percentage points, against the $0.800$\pp\ budget borrowed from another %
paper's worst case. Our own no-removal checkpoints tighten that to $0.698$\pp, and against %
the tighter number the tallest bar here clears it at $0.730$\pp, by a margin far inside that %
budget's own spread. The decisive above-budget cases are elsewhere: two on the second release
and two on MU-Bench CIFAR-10.}
\label{fig:verdict}
\end{figure}

Figure~\ref{fig:verdict} places every crossing against the bound it started from and the
budget it has to clear. Table~\ref{tab:crossings} lists each MU-Bench artifact whose admission status changes
under the rebuild, with its native and rebuilt criterion value. The upper block is
the natively admitted stratum and the lower block the natively excluded one. Removal
ratio is the percentage of the training set the release removed, and seed is the
release's own; the two together with the variant and family identify the released
repository uniquely.

\begin{table}[h]
\caption{The MU-Bench CIFAR-100 artifacts whose published admission status changes when
deployed normalization state is rebuilt from retained data at bit-identical trainable
tensors. Counted on the release's own published cell rather than on our recomputation,
the MU-Bench total across groups is 12 of 14 candidate crossings; this table lists the %
CIFAR-100 block. The benchmark admits an artifact when $|d_f - d_t| \le 0.05$; \emph{out} %
leaves the admitted set and \emph{in} enters it. Release variant is the released
repository prefix, where \texttt{-cl} adds curriculum learning and \texttt{-so} the
release's alternative optimizer.} %
\label{tab:crossings}
\centering\scriptsize
\input{tables/crossings}
\end{table}

The two second-population crossings are SalUn on the deleted class \emph{automobile},
whose retain-train accuracy moves $0.9497 \to 0.9856$ while forget-train accuracy %
moves $0.0018 \to 0.0004$, and SSD on \emph{airplane}, whose retain-train accuracy %
moves $0.8999 \to 0.9951$ with forget-train accuracy at $0.0000$ before and %
after. %
Both cross a threshold of $0.95$ on retain-train accuracy, and neither trades %
forgetting for the pass.

\paragraph{Do the crossings survive independently re-keyed refits?}
A crossing declared on one refit could be an artifact of that refit's retained draw, so we
re-decided every crossing separately on each of the ten refits the census already carries,
under both decision rules, with the rule set fixed beforehand. Table~\ref{tab:xstab} gives
the result. The two CIFAR-10 crossings that clear the measured budget are unanimous on all
ten draws under either rule, and that group's per-draw crossing count does not move. The
CIFAR-100 group is weaker and we report it as such: four of its six published-cell crossings
hold on all ten, one on nine and one on six, and its per-draw count wanders between 6 and 9,
so the CIFAR-100 crossing count is itself draw-dependent. Eleven crossing-rule pairs are
unanimous, each at the exact boundary-null floor of $0.00098$, and all eleven clear the %
Benjamini--Yekutieli cutoff of $0.01656$ under arbitrary dependence. Four of the fourteen %
candidate crossings lie outside the ten-draw stratum and keep exactly the strength the
published-cell re-decision gave them; four further crossings that appear on some draws but
were not declared at the frozen draw are reported in the run artifacts and counted nowhere,
under a bar fixed before the draws were read. The decision instrument carries its own
known-answer control: with every shift zeroed it manufactures no crossing at all.

\begin{table}[h]
\caption{\textbf{The two crossings that clear the measured budget are also the two that
survive every re-keyed refit.} Each candidate crossing re-decided independently on all ten
census refits, under the published-cell and recomputed-cell rules. The last column is how
much the population's crossing count itself moves across those ten draws.}
\label{tab:xstab}
\centering\footnotesize
\input{tables/xstab}
\end{table}

\section{Exposure versus estimation}
\label{app:expose}

Section~\ref{sec:mech} reports the arm that would have turned the displacement into a
deletion failure. This appendix gives its design, its bars, and the state-level reading that
decides which kind of null the cell-level one is.

\paragraph{Design.}
Every arm holds the trainable tensors, the estimator, the producer's training transform and
the size of the fitting pool fixed at 8000 records, and varies only which records fill the
pool. Arm~C takes a retain core of $8000 - k$ records plus $k$ from a disjoint retain
reserve. Arm~D replaces those $k$ with removed records at the share a deployment would draw,
$k = \mathrm{round}(8000 \cdot n_{\mathrm{del}} / 50000)$. Arm~F fills all $k$ slots from the
removed set, which at the release's largest removal ratio is most of the pool. C and D differ
in the provenance of $k$ exchangeable records and in nothing else, pool size included, so the
contrast is exchangeability-exact rather than merely controlled. The twenty checkpoints are
selected by code from the frozen ten-draw ranking, the ten largest and ten smallest
displacements, so none is chosen on a number this wave produced. Each carries six re-keyed
draws whose key stream differs from the census's own, which is why the bar comparing the two
waves is distributional; the tight bar is arm~C reproducing the census's own draw-0 forget
cell, which it does to exactly $0.0$\pp\ on all twenty. All 120 checkpoint-draws re-anchor to %
their own published cells before any contrast is read. %

\paragraph{What the cell-level arms return.}
The pooled forget-accuracy contrast is $+0.015$\pp\ at the deployment share, positive on 5 of %
the 6 draws, and $+0.057$\pp\ at the ceiling, positive on 4 of 6. At six draws the exact %
two-sided floor is $0.03125$, so nothing here is certified in either direction; what is %
measured is the size, and it is small. No checkpoint passes $0.5$\pp\ on either contrast, the %
largest per-checkpoint mean is $+0.271$\pp, and the test-accuracy contrasts behave the same %
way. A confirming outcome was reachable on this instrument: the analogous
maximal-provenance arm on the quantization channel does move 3 of 71 checkpoints past their
margin, one of them by more than six points (Appendix~\ref{app:ptq}).

\begin{table}[h]
\caption{\textbf{What tracks a checkpoint's displacement is how far its shipped state has
drifted from any retain-only refit, not whose data that state was fitted on.} Twenty primary
checkpoints, the ten most and ten least displaced in the frozen census, over six re-keyed
draws. Column~3 counts them past a threshold: in (a) the by-construction null band our refit
measured against the omission-retrained references, $0.0019$; in (b) $0.5$\pp. Strata and %
thresholds were fixed before any distance was computed.} %
\label{tab:mech}
\centering\footnotesize\setlength{\tabcolsep}{4pt}
\input{tables/mech}
\end{table}

\paragraph{Which kind of null it is.}
Table~\ref{tab:mech} carries the arithmetic. A cell-level null can mean the state changed and
the audit could not see it, or that the state did not change. We had already met the first kind: on the quantization channel a large
share of activation scales differs with the deleted records in or out while no published cell
budges (Appendix~\ref{app:ptq}), so there the bound is on propagation. Here the fork was
written down before any distance
existed and the aggregation decides it mechanically against the band, not by reading. It
returns the second branch. In the channel-standardized units of Appendix~\ref{app:controls},
the provenance contrasts sit at $0.00061$ and $0.00101$ with 0 and 1 of 20 checkpoints above %
the by-construction null band of $0.0019$, while the distance from a checkpoint's shipped %
state to a retain-only refit is $0.00538$, clears that band on all twenty, and is $5.43$ %
times larger at the median, ranging from $1.82$ to $15.00$ across them. There is no state %
change for the audit to be insensitive to. %

Two qualifications ride with this. Both provenance distances are \emph{upper} bounds on
provenance, because exchanging $k$ records for $k$ others perturbs a finite-sample estimate
even when provenance is irrelevant and no arm here isolates that resampling component; the
bound is therefore conservative in the direction that matters. And the result bounds what the
removed data contributes to a fitted state at these releases' removal shares. It says nothing
about the census, the crossings or the documentation leg, which are claims about a
convention's effect on an audit's outputs rather than about whose data the state held.

\section{Forgetting quality: two readouts, one usable}
\label{app:drift}

The main text reports that the convention reaches a forgetting-quality readout on one family
of one release. Here is what each population could and could not answer.

\paragraph{The primary release answers neither way.}
Table~\ref{tab:mia} collects both readouts. The field's standard membership attack is
\emph{unreadable} on these checkpoints rather than null: its cross-validation accuracy has a
median of $0.518$ against a $0.55$ floor fixed before the wave, with $6.9\%$ of arm reads %
above it, because these models barely overfit. No count is certified at any depth. The number %
the field publishes as its privacy metric nevertheless moves under the convention, by $-0.789$ %
pooled and up to $3.633$, and holding the attack frozen puts $-0.754$ of that in the model's %
own scores rather than in the attack re-adapting; an identity check with no convention swap %
returns exactly zero, so the readout adds no variance of its own. The drift readout on the
same population is \emph{voided} by the control its own contract named: the pre-unlearning
base, which carries no removal at all, drifts $0.0642$ along the same axis against the %
unlearned checkpoints' $0.0191$, a ratio of $3.36$ with all 15 of 15 cells above the %
threshold whose breach the contract said would void the leg. This is the generic
recalibration rival firing a third time, and we report the leg as voided rather than as a
result.

\begin{table}[h]
\caption{\textbf{Neither forgetting-quality readout on the primary release can answer, and
each fails for its own measured reason.} The membership attack is at chance on checkpoints
that barely overfit; the drift readout is voided by a no-removal control that moves further
than the checkpoints under test.}
\label{tab:mia}
\centering\footnotesize
\input{tables/mia}
\end{table}

\paragraph{The second release ships the oracle the question needs.}
It publishes an omission-retrained reference for every deleted class, so a forgetting position
can be read against a real pole rather than an attack. We measure where the refit moves each
checkpoint along the base-to-oracle axis, over three re-keyed draws, under both pole
conventions. Table~\ref{tab:drift} gives every family. SSD moves $+0.0167$ back toward the %
un-unlearned base on 10 of 10 classes, with the sign holding on all three draws and under the %
alternative pole convention at $+0.0172$; 5 of the 10 classes pass the materiality bar fixed %
in advance. The effect concentrates on shippable checkpoints: $+0.0215$ over the 7 classes %
clearing the release's own retain bar, all 7 same-signed, against $+0.0054$ over the 3 that %
fail it. That split was declared after the first draw and before the other two were read, and %
they confirmed it. The across-draw spread is $0.000138$ against an across-class spread of %
$0.0119$, so the classes disagree far more than the draws do. The other three families sit at %
noise and their signs do not hold across draws.

\begin{table}[h]
\caption{\textbf{One family of four moves on a forgetting-quality readout measured against a
real omission-retrained oracle, and the family that moved its weights furthest is not it.}
Position shift along the base-to-oracle axis under the refit, by family, with the
weight-movement rival read on the same checkpoints. Positive is toward the un-unlearned base.}
\label{tab:drift}
\centering\footnotesize
\input{tables/drift}
\end{table}

\paragraph{The cheapest rival, asked on the same checkpoints.}
That SSD simply moved its weights furthest is the obvious alternative, and it runs against
itself on two axes. Between families the orderings are discordant: SalUn's median relative
weight distance is $0.577$ against SSD's $0.245$, while its drift is $+0.0058$ against SSD's %
$+0.0167$. Within %
SSD the correlation between weight movement and drift is significantly \emph{negative} where %
the account predicts positive, at $\rho = -0.794$ over ten classes with a permutation %
$p = 0.0085$, so the classes whose weights moved furthest drift least. The rule we fixed for %
this stops short of refuted rather than reaching it, because its refutation clause required
the further-moving family to sit at noise and SalUn's drift rests on the 2 of 10 classes its
own release admits. What distinguishes SSD's procedure is therefore open, and the same rival
came back mixed on the other population.

\section{Negative and unresolved results}
\label{app:negatives}

Three questions we asked returned answers that support no claim in the main text. We
report them because a reader deciding what to build on this needs to know which parts
did not hold.

\paragraph{The removed-data provenance channel is not established.}
We asked whether the deployed state alone carries a contrast between removed and
unseen items beyond a matched permutation control. The separation we pre-registered
for it allows at most 1 of 8 artifacts to also reject under a swap from one
retain-only state to a disjoint retain-only state. Two retain-only states carry no
removal-data difference by construction. Four of eight reject, all at $p = 0.000$, so the bar %
failed and no provenance claim rests on this channel. What survives is a magnitude
statement we did not pre-register: across those eight artifacts the
shipped-to-retain-only contrast averages $0.342$ in absolute value and reaches %
$0.6351$, against $0.011$ for retain-only-to-retain-only on the same artifacts, a %
factor of $30.5$. %
That magnitude tracks each artifact's own removed-versus-retained margin gap at a
correlation of $-0.999$, exactly as the rival predicts, so a nuisance-free reading is %
unavailable. %
A rejection count was the wrong instrument at this sample size, where a nuisance
thirty times below the effect still rejects; comparing magnitudes is what the metric
channel's control got right. The readout itself is sound: on a checkpoint that never
underwent removal the permutation $p$ is at worst $0.402$ over the five removal %
ratios. %

\paragraph{The obvious mechanism fails on the family it needs to explain.}
If a rebuilt statistic were simply the shipped one with the removed data's
contribution taken out, the displacement of an artifact would be proportional to the
fraction of the training set removed, with a zero intercept. We pre-declared three
bars for that identity and it passed none of them on the only family that moves.
The within-family correlation between shift and removed fraction is $-0.211$ at %
one-sided $p = 0.81$; leave-one-ratio-out prediction from the through-origin fit %
loses to predicting the family mean; and the ordinary-least-squares intercept
excludes zero on all three published metrics. %
The record therefore establishes that the shift is method-indexed without explaining
what inside a procedure produces it.

We record one contrast this test surfaced, explicitly as an observation and not as a
result, because it was not pre-registered and nothing rests on it. The two families
that are equivalent to zero fit the identity closely and out of sample. Their
correlations run $+0.56$ to $+0.58$ across the three metrics, their intercepts are %
$-0.031$, $+0.131$ and $+0.073$ with every interval containing zero, and %
leave-one-ratio-out beats the mean on all three. %
A probe-size explanation is excluded, since the retained and test probes are fixed
size while only the removed probe scales with the ratio, and the pattern appears
identically on all three. Promoting this to a claim would need a pre-declared
confirmation loop we have not run, and the second population cannot supply one
because class deletion has no removed-fraction axis.

\paragraph{Native accuracy does not explain the census, within family either.}
Across the census the correlation between a checkpoint's displacement on forget accuracy
and its own native accuracy is $-0.137$ on the primary population and $-0.010$ on the %
CIFAR-10 groups. We report the within-family structure rather than suppress it: it reaches %
$-0.699$ on the smallest primary family, at $n=8$, and about $-0.44$ inside each of the two %
large CIFAR-10 families, while the two large primary families are near zero. None of that %
recovers the headroom reading, whose prediction is a pooled negative correlation of the
size of the effect.

\paragraph{No checkpoint-readable predictor transfers across families, and one works
inside a release group.}
The relative distance between an artifact's shipped normalization buffers and the
released base's is computable from two files with no data and no forward pass, and
pooled across families it correlates with the shift. That is entirely a between-family
difference. Within each family the correlations are inconsistent in sign, and
leave-one-family-out linear prediction beats predicting the training mean only
trivially while its errors are as large as the effect being predicted. The number of
batches each layer records having tracked is readable but near-constant across
artifacts and carries no discriminative information.

The third predictor is the strongest rival to reading the displacement as a property of
the artifact, and it is the one we ran last. If a shipped state is simply stale with
respect to weights that moved, then how far the weights moved should predict how far the
cell moves. The relative $L_2$ distance of an artifact's trainable parameters from its own
group's released base is readable from two files. We fixed a three-way decision rule before
computing any correlation: uphold if the within-family rank correlation is significantly
positive in a majority of families, reject if in none, and call it mixed otherwise.
Table~\ref{tab:stale} gives the outcome, which is mixed. The correlation is significantly
positive in 4 of 9 families, all four on the CIFAR-10 groups and reaching $\rho = +0.770$, %
and in none of the primary population's three. Pooled across all 221 checkpoints it is %
$+0.092$ at $p = 0.170$, which does not decide either way, since pooling across families %
whose means differ is a Simpson confound and we declared in advance that it would not.

We report the partial success rather than the headline. Inside the CIFAR-10 groups weight
distance is genuinely informative, clearing significance on all three of that group's
families, so an auditor working at that architecture and dataset could triage on it. The
same auditor would be wrong to carry the screen to the primary CIFAR-100 population, where
it recovers nothing. That is the claim this record supports: no tested predictor works
universally, not that per-artifact measurement is unavoidable. Directional, layerwise and
normalization-layer-restricted weight accounts are untested and one of them could yet
succeed. The reason to run the test per artifact is therefore the within-family dispersion
itself rather than the predictors failing: the family whose interval is equivalent to zero
still contains 8 of its 45 checkpoints displaced past the same margin. %

\begin{table}[h]
\caption{The weight-movement rival, under a rule fixed before any correlation was computed.
Each row is one family inside one release group; $\rho$ is the Spearman correlation between
an artifact's relative weight distance from its own base and its mean forget-accuracy
displacement over ten draws, and $p$ is a within-family permutation test. Four of nine are
significantly positive, all on the CIFAR-10 groups. The pooled row is reported for
completeness and decides nothing, as the pre-registration states.}
\label{tab:stale}
\centering\footnotesize\setlength{\tabcolsep}{4pt}
\input{tables/stale}
\end{table}

\section{Provenance and timeline}
\label{app:timeline}

The operation this paper measures is not new here, and the dates on the public record fix the
order plainly. The recalibration diagnostic appears in the released code accompanying
\citet{purge2026} as \texttt{experiments/bn\_recalibration.py}, entering that repository at the
commit of 24 June 2026 and indexed in its own README as a BN-recalibration attack diagnostic.
The same paper's arXiv v1, dated 2 June 2026, contains no discussion of it: a text search of
that version returns no occurrence of recalibration or of the normalization-free remedy the code
implements. The instrument, bars and decision rules used here were fixed against that
background and before the measurements they govern. The primary population's protocol,
equivalence margin and blocking bars were pre-registered on 29 August 2026 in
\texttt{PREREGISTRATION.md} for the primary wave. The census rows were minted from 30 August
2026 onward. The second deployed-state channel, its operating point and its detection floor were
pre-registered on 6 September 2026 in that arm's own \texttt{PREREGISTRATION.md}. That was before
any displacement on an unlearned artifact in that channel existed. \citet{bnillusion2026} was posted
on 8 September 2026, after both pre-registrations and after the census rows this paper reports.
Every pre-registration file named here is included verbatim in the supplementary code archive, so
a reader can check each bar against the wave it governed rather than take the ordering on trust.

%% file: tables/second.tex
\begin{tabular}{@{}lrccc@{}}
\toprule
Arm & $n$ & \makecell{Retain-class accuracy\\shift (pp, mean $\pm$ sd)} & \makecell{90\% interval\\and verdict} & \makecell{Paired against\\the reference (pp)} \\
\midrule
omission retraining (ref.) & 10 & $+0.014 \pm 0.072$ & [$-0.03$, $+0.06$]~\textsc{e} & -- \\
CF-$k$ & 10 & $+0.053 \pm 0.074$ & [$+0.01$, $+0.10$]~\textsc{e} & $+0.039$ ($p{=}0.247$) \\
AdvNegGrad & 10 & $+1.916 \pm 0.732$ & [$+1.49$, $+2.34$]~\textsc{d} & $+1.901$ ($p < 10^{-4}$) \\
SalUn & 2 & $+2.456 \pm 1.037$ & -- & -- \\
SSD & 10 & $+11.382 \pm 20.138$ & -- & $+1.516$ ($p{=}0.125$) \\
\bottomrule
\end{tabular}

%% file: tables/design.tex
\begin{tabular}{@{}lllcl@{}}\toprule
release, data & backbone & stratum & \makecell{anchored /\\ in census} & carries \\\midrule
MU-Bench, CIFAR-100 & ResNet-50 & admitted & 71 / 71 & the census; family verdicts \\
MU-Bench, CIFAR-10 & ResNet-50/-34 & admitted & 150 / 150 & the census; crossings \\
Unlearning Comparator, & ResNet-18 & all arms & 42 of 50 / -- & the paired contrast \\
\quad CIFAR-10 & & & & against the reference \\
\midrule
\emph{all anchored} & & & 263 / 221 & \\
\midrule
MU-Bench, CIFAR-100 & ResNet-50 & excluded & 42 of 73 & the inward crossings only, \\
\quad \emph{outside both totals} & & & & no census and no anchored total \\
\midrule
MU-Bench, CIFAR-100 & ResNet-50, & quantized & 118 / 118 & the second deployed state \\
\quad \emph{second channel} & swin-base & & & (Sec.~\ref{sec:ptq}); 47 ship no \\
 & & & & normalization state at all \\
\bottomrule\end{tabular}

%% file: tables/census.tex
\begin{tabular}{@{}llrrl@{}}
\toprule
Population stratum & Family & \makecell[r]{Artifacts\\measured} & \makecell[r]{Reproducing all three\\published cells} & Note \\
\midrule
MU-Bench CIFAR-100, admitted & all & 72 & 71 & -- \\
MU-Bench CIFAR-100, excluded & bad-teaching & 22 & 22 & -- \\
MU-Bench CIFAR-100, excluded & random-label & 21 & 20 & -- \\
MU-Bench CIFAR-100, excluded & salun & 30 & 0 & forget cell only (30 of 30) \\
MU-Bench CIFAR-10 groups & all & 205 & 150 & 55 SalUn fail, 54 on the forget cell alone \\
Unlearning Comparator & all & 50 & 42 & 8 SalUn artifacts excluded \\
\bottomrule
\end{tabular}

%% file: tables/percensus.tex
\begin{tabular}{@{}llccccl@{}}\toprule
group & family & $n$ & forget & retain & test & \makecell[l]{family verdict,\\forget accuracy} \\\midrule
CIFAR-100 ResNet-50 (8 rep.) & all & 71 & 20 & 16 & 15 & \\
 & bad teaching & 18 & 12 & 10 & 10 & \textsc{unresolved} \\
 & neggrad & 45 & 8 & 6 & 5 & \textsc{equivalent} \\
 & random label & 8 & 0 & 0 & 0 & \textsc{equivalent} \\
\addlinespace
CIFAR-10 ResNet-50/34 (8 rep.) & all & 150 & 28 & 23 & 25 & \\
 & bad teaching & 40 & 8 & 7 & 8 & -- \\
 & neggrad & 55 & 14 & 13 & 13 & -- \\
 & random label & 55 & 6 & 3 & 4 & -- \\
\addlinespace
\midrule
both groups, 8 replicates & & 221 & 48 & 39 & 40 & \\
both groups, 10 replicates & & 221 & 47 & 39 & 40 & \\
\bottomrule\end{tabular}

%% file: tables/sens.tex
\begin{tabular}{@{}lccc@{}}\toprule
\multicolumn{4}{@{}l}{\textbf{(a) the margin.} displaced checkpoints of 221, at the sign-test floor / certified} \\
margin & $d_f$ & $d_r$ & $d_t$ \\\midrule
$\pm 1.0$\pp & 56 / 56 & 48 / 48 & 47 / 47 \\
$\pm 1.2$\pp & 47 / 47 & 39 / 39 & 40 / 40 \\
$\pm 1.5$\pp & 31 / 31 & 28 / 28 & 30 / 30 \\
$\pm 2.0$\pp & 23 / 0 & 21 / 0 & 19 / 0 \\
\midrule
\multicolumn{4}{@{}l}{\textbf{(b) the fitting pool.} 71 primary-population checkpoints, one draw each} \\
& mean $d_f$ & past & $|\Delta|$ vs.\ 8000 \\
examples & shift (\pp) & margin & median / max \\\midrule
4000 & $+1.181$ & 25 & $0.150$ / $0.600$ \\
8000 & $+1.191$ & 26 & -- \\
16000 & $+1.190$ & 25 & $0.100$ / $0.300$ \\
\bottomrule\end{tabular}

%% file: tables/controls.tex
\begin{tabular}{@{}p{2.5cm} p{3.2cm} p{7.2cm}@{}}
\toprule
Control & What it rules out & Measurement \\
\midrule
Choice of retained subset & the rebuild itself, with no removal-data difference & $0.181$ pp between two disjoint retain-only pools, against $2.558$ pp for shipped~$\to$~retain-only ($14.2\times$, 68 of 71 artifacts) \\
Accumulation rule & the estimator's functional form & on a no-removal checkpoint, test accuracy moves $-2.550$ pp (exact) and $-2.583$ pp (EMA) with the transform held fixed \\
Data distribution & a transform mismatch inside the rebuild & the same checkpoint moves $-0.077$ pp under the producer's training transform and $-2.550$ pp under its evaluation transform \\
Native accuracy level & headroom rather than the producing method & level-matched, the moving family gains $1.614$ pp more forget accuracy ($p{=}0.0052$) while sitting $6.4$ pp \emph{higher} natively \\
Membership readout & a contrast manufactured by the readout itself & on a checkpoint that never underwent removal the permutation $p$ is at worst $0.402$ over the five removal ratios \\
Scoring pipeline & a scorer that reports structure where there is none & under permuted labels the pipelines return $0.0082$ and $0.1035$ against chance rates of $0.01$ and $0.1$ \\
\bottomrule
\end{tabular}

%% file: tables/zrf.tex
\begin{tabular}{@{}lccccc@{}}\toprule
family & $n$ & mean shift & median $|$shift$|$ & displaced ($M$) & displaced ($M_{\mathrm{s}}$) \\\midrule
bad teaching & 18 & $+0.0115$ & $0.0082$ & 18 & 14 \\
neggrad & 45 & $-0.0012$ & $0.0014$ & 3 & 2 \\
random label & 9 & $-0.0009$ & $0.0005$ & 0 & 0 \\
\midrule
all three families & 72 & & & 21 & 16 \\
no-removal base checkpoint & 1 & $-0.0007$ & & 0 & 0 \\
\bottomrule\end{tabular}

%% file: tables/ptq.tex
\begin{tabular}{@{}lcc@{}}\toprule
& ResNet-50 & swin-base \\
& \emph{batch norm} & \emph{layer norm} \\\midrule
anchored checkpoints & 71 & 47 \\
quantization sites per checkpoint & 52 & 149 \\
\midrule
displaced past the $1.2$\pp\ margin ($d_f$/$d_r$/$d_t$) & 0/0/0 & 0/0/0 \\
consistent-sign displacement at margin $0$ ($d_f$) & 0 & 0 \\
same, nuisance contrast that moves no removed record & 0 & 0 \\
largest mean displacement (\pp, $d_f$) & $0.287$ & $0.169$ \\
no-removal base checkpoint, $|$mean$|$ (\pp, $d_f$) & $0.048$ & $0.042$ \\
\midrule
detection floor, consistent-sign test (90\%) & $1.0$\pp & $0.5$\pp \\
detection floor, displacement census (90\%) & $2.0$\pp & $2.0$\pp \\
scales that differ with the removed records & $12.1$\% & $11.2$\% \\
\midrule
ceiling arm, checkpoints past the margin & 3 of 71 & 0 of 47 \\
ceiling arm, largest displacement (\pp) & $6.14$ & $0.27$ \\
\bottomrule\end{tabular}

%% file: tables/gen.tex
\begin{tabular}{@{}llrrrrrr@{}}\toprule
& forget-set & & admis- & native & \multicolumn{2}{c}{mean $d_f$ shift (\pp)} & dis- \\
cell & exposure & $n$ & sible & $d_t$ & all & admissible & placed \\\midrule
no-removal base & -- & 5 & 5 & $95.7$ & $-0.010$ & $-0.010$ & 0 \\
omission retrain & none & 5 & 5 & $95.5$ & $+0.024$ & $+0.024$ & 0 \\
retain-only fine-tune & none & 5 & 5 & $95.7$ & $-0.006$ & $-0.006$ & 0 \\
retain-only fine-tune, frozen & none & 5 & 0 & $86.2$ & $+11.119$ & -- & 0 \\
gradient ascent & pure forget & 5 & 1 & $24.5$ & $+0.000$ & $+0.000$ & 0 \\
gradient ascent, frozen & pure forget & 5 & 1 & $24.5$ & $+0.004$ & $+0.022$ & 0 \\
masked relabelling & mixed & 5 & 4 & $88.8$ & $+3.183$ & $+0.473$ & 2 \\
masked relabelling, frozen & mixed & 5 & 0 & $13.0$ & $+79.614$ & -- & 0 \\
\bottomrule\end{tabular}

%% file: tables/genfacts.tex
\begin{tabular}{@{}lr@{}}\toprule
base checkpoints, mean test accuracy over 5 seeds & $95.697$ \\
per-cell margin, $d_f$ / $d_r$ / $d_t$ (\pp) & $0.323$ / $0.005$ / $0.044$ \\
artifacts, all draws present & 40 of 40 \\
admitted by the gap criterion alone & 40 of 40 \\
of those, below their own retrain reference & 19 \\
admitted once a utility floor is added & 21 \\
distinct refit buffer hashes / weight hashes per artifact & 10 / 1 \\
artifacts returning one $d_f$ value across all ten draws & 18 \\
forget-metric quantum (\pp) & $0.022$ \\
frozen twin, weight distance ratio to its default sibling & $1.0$ \\
\bottomrule\end{tabular}

%% file: tables/reach.tex
\begin{tabular}{@{}lrrrr@{}}\toprule
\multicolumn{5}{@{}l}{\textbf{A. per-checkpoint census, ten re-keyed refits}} \\
metric & $n$ & margin (pp) & displaced & median sd (pp) \\\midrule
forget accuracy & 45 & $1.868$ & 20 & $0.099$ \\
retained accuracy & 45 & $1.292$ & 20 & $0.096$ \\
test accuracy & 45 & $0.803$ & 22 & $0.070$ \\
\bottomrule\end{tabular}

\vspace{4pt}

\begin{tabular}{@{}p{86mm}r@{}}\toprule
\multicolumn{2}{@{}l}{\textbf{B. what a verdict crossing would have taken here}} \\\midrule
recalibration budget, from this population's own five bases & $0.085$\pp \\
checkpoints whose verdict crosses, default policy & 0 of 45 \\
checkpoints the rebuild moves \emph{away} from the bound & 33 of 45 \\
checkpoints above the bound, where only a narrowing converts & 30 of 45 \\
smallest gap shift this stratum could have converted & $2.361$\pp \\
largest gap shift observed & $3.003$\pp \\
by how much the nearest checkpoint missed & $0.204$\pp \\
\bottomrule\end{tabular}

%% file: tables/ladder.tex
\begin{tabular}{@{}lrrrr@{}}\toprule
\multicolumn{5}{@{}l}{\textbf{A. pairs matched on weight distance alone}} \\
rungs & $|\ln$ ratio$|$ & frozen twin admissible & twin contrast (pp) & sd \\\midrule
R1\,/\,R0 & $0.117$ & 5 of 5 & $+0.07$ & $0.21$ \\
R2\,/\,R1 & $0.055$ & 3 of 5 & $+13.04$ & $28.07$ \\
R5\,/\,R3 & $0.092$ & 0 of 5 & $+51.37$ & $25.71$ \\
\bottomrule\end{tabular}

\vspace{4pt}

\begin{tabular}{@{}p{86mm}r@{}}\toprule
\multicolumn{2}{@{}l}{\textbf{B. the curve reading, and what the instrument can see}} \\\midrule
detection floor for a twin contrast on this forget set & $12.93$\pp \\
exposed arm above the unexposed arm's own distance curve & $+33.80$\pp \\
its interval over 5 seed clusters & $[21.28, 46.32]$ \\
same contrast at the unmatched operating point & $+65.31$\pp \\
its spread, and how many seeds agree in sign & $10.57$, 5 of 5 \\
pairs with identical trainable weights and different buffers & 90 of 90 \\
default rungs against the frozen table of Appendix~\ref{app:gen} & $0.000$\pp \\
\bottomrule\end{tabular}

%% file: tables/bnmode.tex
\begin{tabular}{@{}lllcrrr@{}}\toprule
& & source-read & & & mean & dis- \\
release & family & exposure & from & $n$ & shift (\pp) & placed \\\midrule
MU-Bench CIFAR-100 & bad-teaching & \textsc{high} & read & 18 & $+2.483$ & 12 \\
MU-Bench CIFAR-100 & neggrad & \textsc{high} & read & 45 & $+0.741$ & 8 \\
MU-Bench CIFAR-100 & random-label & \textsc{low} & read & 8 & $+0.362$ & 0 \\
Unlearning Comparator & SalUn & \textsc{high} & read & 2 & $+2.456$ & -- \\
Unlearning Comparator & AdvNegGrad & \textsc{high} & inferred & 10 & $+1.916$ & -- \\
Unlearning Comparator & CF-k & \textsc{zero} & inferred & 10 & $+0.053$ & -- \\
Unlearning Comparator & omission retrain & \textsc{zero} & read & 10 & $+0.014$ & -- \\
\midrule
\multicolumn{7}{@{}l}{\emph{tests fixed before any class met a displacement}} \\
mean shift, \textsc{high} families & \multicolumn{6}{r}{$+1.899$\pp} \\
mean shift, \textsc{zero} families & \multicolumn{6}{r}{$+0.034$\pp} \\
separation, against a $1.2$\pp\ margin & \multicolumn{6}{r}{$1.865$\pp} \\
ordering test, exact $p$ (floor $0.0556$) & \multicolumn{6}{r}{$0.0556$} \\
\bottomrule\end{tabular}

%% file: tables/primary.tex
\begin{tabular}{@{}l cccc@{}}
\toprule
 & \multicolumn{4}{c}{Shift under the retain-only state rebuild (percentage points)} \\
\cmidrule(l){2-5}
Published metric & \makecell{bad-teaching\\$n{=}18$, $G{=}3$\\unresolved} & \makecell{neggrad\\$n{=}45$, $G{=}3$\\equivalent} & \makecell{random-label\\$n{=}8$, $G{=}3$\\equivalent} & \makecell{no-removal base\\$n{=}3$ seeds\\ } \\
\midrule
Forget accuracy & $+2.47$ [$-0.12$, $+5.06$] & $+0.83$ [$+0.49$, $+1.17$] & $+0.35$ [$-0.24$, $+0.94$] & $+0.21$ \\
Retain accuracy & $+2.23$ [$+0.08$, $+4.38$] & $+0.90$ [$+0.63$, $+1.18$] & $+0.05$ [$-0.05$, $+0.15$] & $-0.05$ \\
Test accuracy & $+1.86$ [$-0.26$, $+3.98$] & $+0.84$ [$+0.58$, $+1.10$] & $+0.10$ [$+0.03$, $+0.17$] & $-0.08$ \\
\midrule
\multicolumn{5}{@{}l@{}}{\emph{Half-width returned on a planted true zero at each family's own clustering, forget / retain / test}} \\
\emph{detection floor} & $\pm2.59$/$\pm2.15$/$\pm2.12$ & $\pm0.34$/$\pm0.28$/$\pm0.26$ & $\pm0.59$/$\pm0.10$/$\pm0.07$ & --- \\
\bottomrule
\end{tabular}

%% file: tables/estimators.tex
\begin{tabular}{@{}llccccc@{}}
\toprule
Family & Metric & \makecell{unclustered} & \makecell{CR2 by\\variant} & \makecell{CR2 by variant\\$\times$ ratio} & \makecell{CR2 by\\seed} & \makecell{crossed\\hierarchical} \\
\midrule
bad-teaching & forget & [$+1.74$, $+3.20$]~\textsc{d} & [$-0.12$, $+5.06$]~\textsc{u} & [$+1.43$, $+3.51$]~\textsc{d} & [$+1.91$, $+3.02$]~\textsc{d} & [$+1.11$, $+3.73$]~\textsc{u} \\
 & retain & [$+1.62$, $+2.84$]~\textsc{d} & [$+0.08$, $+4.38$]~\textsc{u} & [$+1.35$, $+3.12$]~\textsc{d} & [$+1.75$, $+2.72$]~\textsc{d} & [$+1.32$, $+3.33$]~\textsc{d} \\
 & test & [$+1.26$, $+2.46$]~\textsc{d} & [$-0.26$, $+3.98$]~\textsc{u} & [$+1.01$, $+2.71$]~\textsc{u} & [$+1.33$, $+2.38$]~\textsc{d} & [$+0.84$, $+2.91$]~\textsc{u} \\
\addlinespace
neggrad & forget & [$+0.65$, $+1.01$]~\textsc{e} & [$+0.49$, $+1.17$]~\textsc{e} & [$+0.56$, $+1.10$]~\textsc{e} & [$+0.61$, $+1.05$]~\textsc{e} & [$+0.40$, $+1.19$]~\textsc{e} \\
 & retain & [$+0.74$, $+1.07$]~\textsc{e} & [$+0.63$, $+1.18$]~\textsc{e} & [$+0.66$, $+1.15$]~\textsc{e} & [$+0.69$, $+1.12$]~\textsc{e} & [$+0.53$, $+1.24$]~\textsc{u} \\
 & test & [$+0.68$, $+0.99$]~\textsc{e} & [$+0.63$, $+1.04$]~\textsc{e} & [$+0.58$, $+1.10$]~\textsc{e} & [$+0.65$, $+1.02$]~\textsc{e} & [$+0.47$, $+1.19$]~\textsc{e} \\
\addlinespace
random-label & forget & [$+0.12$, $+0.58$]~\textsc{e} & [$-0.24$, $+0.94$]~\textsc{e} & [$+0.07$, $+0.63$]~\textsc{e} & [$+0.35$, $+0.35$]~\textsc{e} & [$+0.05$, $+0.62$]~\textsc{e} \\
 & retain & [$-0.03$, $+0.12$]~\textsc{e} & [$-0.02$, $+0.12$]~\textsc{e} & [$-0.05$, $+0.15$]~\textsc{e} & [$-0.30$, $+0.40$]~\textsc{e} & [$-0.24$, $+0.13$]~\textsc{e} \\
 & test & [$-0.01$, $+0.20$]~\textsc{e} & [$+0.09$, $+0.11$]~\textsc{e} & [$+0.03$, $+0.17$]~\textsc{e} & [$-0.28$, $+0.47$]~\textsc{e} & [$-0.00$, $+0.20$]~\textsc{e} \\
\bottomrule
\end{tabular}

%% file: tables/planted.tex
\begin{tabular}{@{}ll cl cl@{}}
\toprule
 & & \multicolumn{2}{c}{planted true effect $+0.0$\pp} & \multicolumn{2}{c}{planted true effect $+3.0$\pp} \\
\cmidrule(lr){3-4}\cmidrule(l){5-6}
Family & Metric & recovered [90\% interval] & verdict & recovered [90\% interval] & verdict \\
\midrule
bad-teaching & forget & $+0.00$ [$-2.59$, $+2.59$] & unresolved & $+3.00$ [$+0.41$, $+5.59$] & unresolved \\
 & retain & $+0.00$ [$-2.15$, $+2.15$] & unresolved & $+3.00$ [$+0.85$, $+5.15$] & unresolved \\
 & test & $+0.00$ [$-2.12$, $+2.12$] & unresolved & $+3.00$ [$+0.88$, $+5.12$] & unresolved \\
\addlinespace
neggrad & forget & $+0.00$ [$-0.34$, $+0.34$] & equivalent & $+3.00$ [$+2.66$, $+3.34$] & different \\
 & retain & $+0.00$ [$-0.28$, $+0.28$] & equivalent & $+3.00$ [$+2.72$, $+3.28$] & different \\
 & test & $+0.00$ [$-0.26$, $+0.26$] & equivalent & $+3.00$ [$+2.74$, $+3.26$] & different \\
\addlinespace
random-label & forget & $+0.00$ [$-0.59$, $+0.59$] & equivalent & $+3.00$ [$+2.41$, $+3.59$] & different \\
 & retain & $+0.00$ [$-0.10$, $+0.10$] & equivalent & $+3.00$ [$+2.90$, $+3.10$] & different \\
 & test & $+0.00$ [$-0.07$, $+0.07$] & equivalent & $+3.00$ [$+2.93$, $+3.07$] & different \\
\bottomrule
\end{tabular}

%% file: tables/superseded.tex
\begin{tabular}{@{}lccc@{}}
\toprule
 & \multicolumn{3}{c}{Evaluation-transform rebuild / distribution-matched rebuild (pp)} \\
\cmidrule(l){2-4}
Published metric & bad-teaching & neggrad & random-label \\
\midrule
Forget accuracy & $+5.022$ / $+2.469$ & $-1.030$ / $+0.830$ & $-0.081$ / $+0.350$ \\
Retain accuracy & $+4.506$ / $+2.234$ & $-1.382$ / $+0.904$ & $-3.541$ / $+0.049$ \\
Test accuracy & $+4.301$ / $+1.859$ & $-0.830$ / $+0.836$ & $-2.099$ / $+0.096$ \\
\bottomrule
\end{tabular}

%% file: tables/verdicts.tex
\begin{tabular}{@{}llcrrl@{}}
\toprule
Release & Success criterion & \makecell{Native\\status} & $n$ & \makecell[r]{Cross-\\ings} & \makecell[l]{Distance past\\the bound} \\
\midrule
MU-Bench & $|d_f - d_t| \le 0.05$ & admitted & 71 & 8 & $0.0002$--$0.0073$ over \\
MU-Bench & $|d_f - d_t| \le 0.05$ & excluded & 42 & 4 & $0.0025$--$0.0063$ under \\
Comparator & $\mathrm{RA} \ge 0.95$, $\mathrm{UA} \le 0.10$ & failing & 4 & 2 & $0.9497{\to}0.9856$, $0.8999{\to}0.9951$ \\
\bottomrule
\end{tabular}

%% file: tables/worked.tex
\begin{tabular}{@{}p{4.0cm} p{4.1cm} p{1.9cm} p{2.4cm}@{}}
\toprule
Released artifact & What the release published & After the rebuild & Consequence \\
\midrule
MU-Bench, bad-teaching, 2\% removed & $|d_f-d_t| = 0.0414$, admitted & $0.0573$ & leaves the criterion \\
MU-Bench, random-label, 8\% removed & $|d_f-d_t| = 0.0503$, not admitted & $0.0437$ & enters the criterion \\
Comparator, SalUn, class \emph{automobile} & $\mathrm{RA} = 0.9497$, fails & $0.9856$ & passes the criterion \\
MU-Bench, salun, all 30 artifacts & forget cell not reproducible under our forget-set convention; retain and test reproduce to $10^{-4}$ & not evaluable on the criterion & excluded, since the criterion is a function of the forget cell \\
\bottomrule
\end{tabular}

%% file: tables/crossings.tex
\begin{tabular}{@{}llrrrrl@{}}
\toprule
\makecell[l]{Release\\variant} & Family & \makecell[r]{Removal\\ratio (\%)} & Seed & \makecell[r]{Native\\$|d_f-d_t|$} & \makecell[r]{Rebuilt\\$|d_f-d_t|$} & \makecell[l]{Crossing\\direction} \\
\midrule
\texttt{unlearn-so} & bad-teaching & 8 & 100 & $0.0404$ & $0.0502$ & out \\
\texttt{unlearn-so} & random-label & 4 & 87 & $0.0496$ & $0.0504$ & out \\
\texttt{unlearn-cl} & random-label & 4 & 87 & $0.0478$ & $0.0505$ & out \\
\texttt{unlearn} & bad-teaching & 2 & 87 & $0.0412$ & $0.0510$ & out \\
\texttt{unlearn} & bad-teaching & 6 & 42 & $0.0407$ & $0.0516$ & out \\
\texttt{unlearn-cl} & bad-teaching & 4 & 87 & $0.0486$ & $0.0530$ & out \\
\texttt{unlearn-so} & bad-teaching & 2 & 21 & $0.0485$ & $0.0547$ & out \\
\texttt{unlearn} & bad-teaching & 2 & 42 & $0.0414$ & $0.0573$ & out \\
\midrule
\texttt{unlearn} & random-label & 8 & 42 & $0.0504$ & $0.0475$ & in \\
\texttt{unlearn-so} & bad-teaching & 8 & 13 & $0.0528$ & $0.0474$ & in \\
\texttt{unlearn} & bad-teaching & 4 & 87 & $0.0508$ & $0.0465$ & in \\
\texttt{unlearn} & random-label & 8 & 87 & $0.0503$ & $0.0437$ & in \\
\bottomrule
\end{tabular}

%% file: tables/xstab.tex
\begin{tabular}{@{}llrrrr@{}}\toprule
& decided on & frozen & on 10 & on 1--9 & per-draw \\
population & the & crossings & of 10 & of 10 & count \\\midrule
MU-Bench CIFAR-100 & published cell & 6 & 4 & 2 & 6--9 \\
MU-Bench CIFAR-100 & recomputed cell & 8 & 3 & 5 & 6--10 \\
MU-Bench CIFAR-10 & published cell & 2 & 2 & 0 & 2--2 \\
MU-Bench CIFAR-10 & recomputed cell & 2 & 2 & 0 & 2--3 \\
\midrule
\multicolumn{4}{@{}l}{crossing--rule pairs unanimous on all 10 draws} & \multicolumn{2}{r}{11} \\
\multicolumn{4}{@{}l}{exact floor $p$ per pair, and the Benjamini--Yekutieli cutoff} & \multicolumn{2}{r}{$0.00098$ / $0.01656$} \\
\multicolumn{4}{@{}l}{crossings outside the ten-draw stratum, left at their own strength} & \multicolumn{2}{r}{4 of 14} \\
\bottomrule\end{tabular}

%% file: tables/mech.tex
\begin{tabular}{@{}lrrrr@{}}\toprule
& & above & \multicolumn{2}{c}{by census stratum} \\
what the arm pair varies & median & band & 10 largest & 10 smallest \\\midrule
\multicolumn{5}{@{}l}{\emph{(a) the deployed state itself, in units of a channel's own activation spread}} \\
shipped $\to$ retain-only refit & $0.00538$ & 20 & $0.00807$ & $0.00287$ \\
provenance of $k$, deployment share & $0.00061$ & 0 & $0.00061$ & $0.00061$ \\
provenance of $k$, every removed record & $0.00101$ & 1 & -- & -- \\
\addlinespace
\multicolumn{5}{@{}l}{\emph{(b) the published forget cell the same arms produce (\pp), pooled over 6 re-keyed draws}} \\
provenance of $k$, deployment share & $+0.015$ & 0 & $+0.006$ & $+0.025$ \\
provenance of $k$, every removed record & $+0.057$ & 0 & $+0.102$ & $+0.012$ \\
\bottomrule\end{tabular}

%% file: tables/mia.tex
\begin{tabular}{@{}llr@{}}\toprule
quantity & & value \\\midrule
\multicolumn{3}{@{}l}{\emph{membership readout: unreadable, not null}} \\
attack cross-validation accuracy, median & & $0.5179$ \\
floor fixed before the wave, and the fraction above it & & $0.55$ / $0.069$ \\
the privacy cell moves anyway, pooled and at its largest (\pp) & & $-0.789$ / $3.633$ \\
of which the model's own scores, attack held frozen (\pp) & & $-0.754$ \\
identity check: no convention swap, no attack-seed effect (\pp) & & $0.0$ \\
\addlinespace
\multicolumn{3}{@{}l}{\emph{drift readout: voided by its own no-removal control}} \\
unlearned artifacts drift, mean & & $0.0191$ \\
the pre-unlearning base, carrying no removal, drifts & & $0.0642$ \\
ratio, against a contract that voids the leg above $0.02$ & & $3.36$ \\
base cells above that threshold & & 15 of 15 \\
\bottomrule\end{tabular}

%% file: tables/drift.tex
\begin{tabular}{@{}lrrrrrrr@{}}\toprule
& mean & $+$ on & past & & weight & \multicolumn{2}{c}{within family} \\
family & $\Delta\lambda$ & all & $0.02$ & $n$ & moved & $\rho$ & perm.\ $p$ \\\midrule
SSD & $+0.0167$ & 10 & 5 & 10 & $0.245$ & $-0.794$ & $0.0085$ \\
SalUn & $+0.0058$ & 2 & 0 & 2 & $0.577$ & -- & -- \\
AdvNegGrad & $+0.0000$ & 4 & 0 & 10 & $0.182$ & $-0.176$ & $0.6318$ \\
CF-k & $-0.0001$ & 2 & 0 & 10 & $0.093$ & $-0.515$ & $0.1339$ \\
\midrule
\multicolumn{6}{@{}l}{SSD, the 7 classes clearing the release's own retain bar} & \multicolumn{2}{r}{$+0.0215$} \\
\multicolumn{6}{@{}l}{SSD, the 3 classes failing it} & \multicolumn{2}{r}{$+0.0054$} \\
\multicolumn{6}{@{}l}{SSD, spread across draws and across classes} & \multicolumn{2}{r}{$0.000138$ / $0.0119$} \\
\bottomrule\end{tabular}

%% file: tables/stale.tex
\begin{tabular}{@{}llrrrrr@{}}\toprule
& & & mean rel. & mean $d_f$ & Spearman & perm. \\
group & family & $n$ & $L_2$ & shift (\pp) & $\rho$ & $p$ \\\midrule
CIFAR-10 ResNet-34 & bad-teaching & 10 & $0.029$ & $+0.681$ & $-0.152$ & $0.682$ \\
CIFAR-10 ResNet-34 & neggrad & 10 & $0.006$ & $+0.738$ & $+0.770$ & $0.014$ \\
CIFAR-10 ResNet-34 & random-label & 10 & $0.840$ & $+1.694$ & $-0.018$ & $0.973$ \\
CIFAR-10 ResNet-50 & bad-teaching & 30 & $0.057$ & $+2.221$ & $+0.528$ & $0.003$ \\
CIFAR-10 ResNet-50 & neggrad & 45 & $0.004$ & $+0.711$ & $+0.431$ & $0.003$ \\
CIFAR-10 ResNet-50 & random-label & 45 & $0.260$ & $+0.391$ & $+0.350$ & $0.021$ \\
CIFAR-100 ResNet-50 & bad-teaching & 18 & $0.016$ & $+2.483$ & $-0.344$ & $0.164$ \\
CIFAR-100 ResNet-50 & neggrad & 45 & $0.004$ & $+0.741$ & $+0.156$ & $0.305$ \\
CIFAR-100 ResNet-50 & random-label & 8 & $0.038$ & $+0.362$ & $+0.595$ & $0.133$ \\
\midrule
pooled across all 221 checkpoints & & 221 & & & $+0.092$ & $0.170$ \\
\bottomrule\end{tabular}